%% file: main.tex
\documentclass{article} 
\usepackage{iclr2025_conference,times}

\input{math_commands.tex}

\usepackage{hyperref}
\usepackage{url}
\usepackage{microtype}
\usepackage{graphicx}
\usepackage{subcaption}
\usepackage{booktabs} 
\usepackage{graphicx}  
\usepackage{lipsum} 
\usepackage{listings}
\usepackage{multirow}
\usepackage{float} 
\usepackage[breakable]{tcolorbox}

\usepackage{hyperref}

\title{DeduCE: Deductive Consistency as a Framework to Evaluate LLM Reasoning}

\author{Atharva Pandey\thanks{Equal Contribution} , Kshitij Dubey$^*$, Rahul Sharma \& Amit Sharma   \\
Microsoft Research India\\
\texttt{\{t-atpandey, t-ksdubey, rahsha, amshar\}@microsoft.com} \\
}

\tcbuselibrary{listings, skins}
\usepackage{fancyhdr}
\iclrfinalcopy 

\begin{document}

\maketitle

\begin{abstract}
Despite great performance on Olympiad-level reasoning problems, frontier large language models can still struggle on high school math when presented with \textit{novel} problems outside standard benchmarks. Going beyond final accuracy, we propose a \textit{deductive consistency} metric to analyze  chain-of-thought output from language models (LMs).Formally, deductive reasoning involves two subtasks: understanding a set of input premises and inferring the conclusions that follow from them. The proposed metric studies LMs' performance on these subtasks, with the goal of explaining LMs' reasoning errors on novel problems: how well do LMs understand input premises with increasing  context lengths, and how well can they infer conclusions over multiple reasoning hops? Since existing benchmarks may be memorized, we develop a pipeline to evaluate LMs' deductive consistency on novel, perturbed versions of benchmark problems. On novel grade school math problems (GSM-8k), we find that LMs are fairly robust to increasing number of input premises, but suffer significant accuracy decay as the number of reasoning hops is increased. Interestingly, these errors are masked in the original benchmark as all models achieve near 100\% accuracy.  As we increase the number of solution steps using a synthetic dataset, prediction over multiple hops still remains the major source of error compared to understanding input premises. Other factors, such as shifts in language style or natural propagation of early errors do not explain the trends. 
 Our analysis provides a new view to characterize LM reasoning---as computations over a window of input premises and reasoning hops---that can provide unified evaluation across problem domains.
\end{abstract}

\section{Introduction}
\label{intro}

Chain-of-thought prompting, the practice of instructing a language model (LM) to output its intermediate steps before the final answer, has led to significant gains on reasoning tasks such as math~\citep{10.5555/3600270.3602070}, logic~\citep{PrOntoQA,PrOntoQAOOD,parmar2024towards}, and language tasks~\citep{suzgun2022challenging}. Recent work shows that models can also solve Olympiad-level problems~\citep{gao2024omnimathuniversalolympiadlevel}.  
However, a parallel stream of work shows that LLMs are sensitive to simple perturbations of the original question, such as changing the numeric values occurring in grade school word problems~\citep{gsm-symbolic,srivastava2024functional}. Importantly, these perturbation do not change the difficulty level of a problem, yet accuracy of frontier LLMs such as GPT-4 significantly reduces. Other studies show a similar \textit{reasoning gap} between original and perturbed problems covering math, logic~\citep{wu2024reasoningreciting} and syllogisms~\citep{lewis2024using}, but why this gap arises is less explored.

\begin{figure*}[th]
    \centering
    \includegraphics[width=1\textwidth]{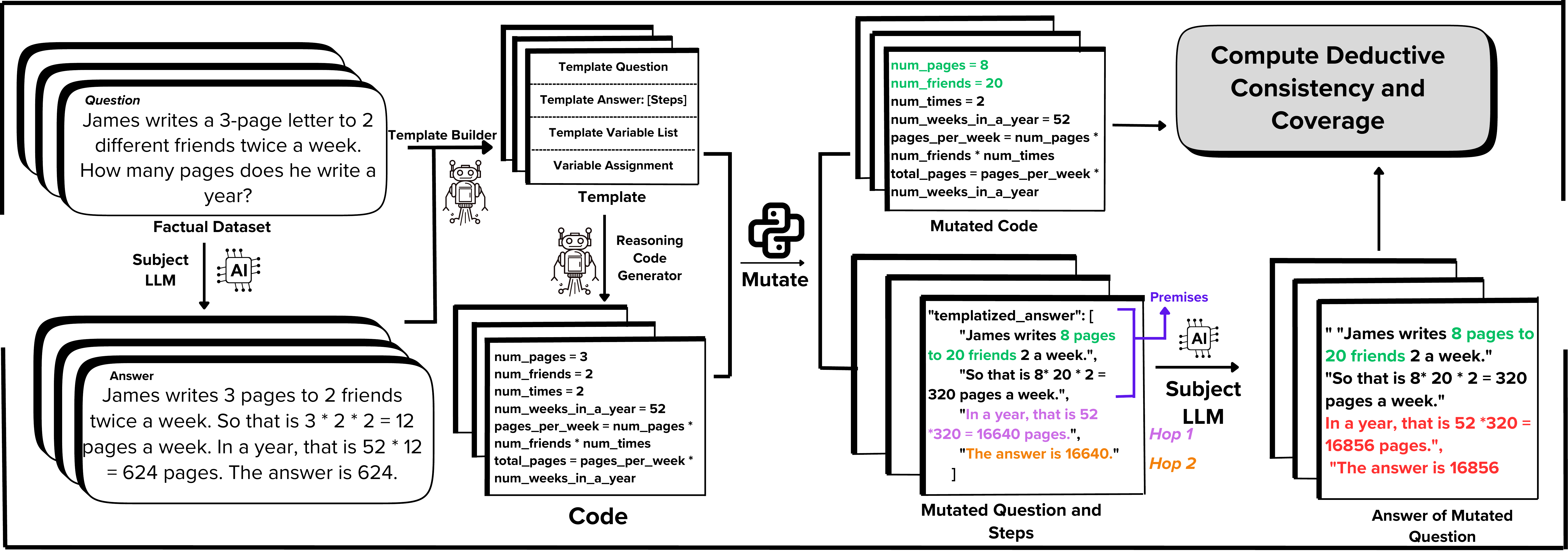}  
    \caption{Pipeline for Deductive Consistency Evaluation. Our method requires access to a single valid solution to compute deductive consistency across multiple premises and reasoning hops. Given a benchmark problem, we use pre-trained LMs to templatize its solution and obtain an executable code solution. Then we update the variables' values in the code and generate a novel problem on which the subject LM is evaluated. For any (number of premises, number of hops) combination, we assess whether the subject LM's solution contains the correct values of the variables. }
    \label{fig:pipeline}
\end{figure*}
In this work, we analyze the reasoning capabilities of language models (LMs) by assessing their consistency with an ideal deductive reasoner. To this end, we introduce \textit{deductive consistency}, a metric that evaluates an AI system’s reasoning validity based on its intermediate steps. We conceptualize a reasoning problem as comprising a set of premises and a target predicate or variable. Given a set of inference rules, the goal of a reasoning system is to determine whether the target predicate can be derived from the input premises (or infer the value of the target variable).
An ideal deductive reasoner operates by systematically deriving the correct reasoning steps from \textit{any} given set of premises, eventually reaching the target predicate if it is logically deducible. Our key insight is to consider the chain-of-thought trace of an LM as a deductive proof and construct \textit{partial} traces to evaluate against an ideal reasoner. Specifically, for a given reasoning problem, we provide an LM with the initial premises (corresponding to problem definition) and append the first few correct reasoning steps before allowing it to continue its generation. This approach enables us to measure consistency with an ideal deductive reasoner across different input premises and reasoning hops, offering a more fine-grained evaluation of LM reasoning beyond final accuracy.

For real world problems, however, obtaining an ideal reasoner is a key challenge. In addition, an LM may have memorized existing benchmark problems, thus making it difficult to evaluate deductive consistency on existing benchmarks. We find a solution by evaluating LM's reasoning on novel, perturbed versions of benchmark problems while utilizing the memorization capabilities of recent LMs to obtain an ideal reasoner based on the original benchmark problem.
Specifically, as Figure~\ref{fig:pipeline} shows, the pipeline first produces a chain-of-thought solution for the original benchmark problem, then translates it into a general solution (python code). We apply multiple correctness checks to filter out original problems that lack a corresponding code representation. For each novel problem, we update the Python code to reflect the new variable values, then convert the modified code back into natural language to generate the correct solution steps, which in turn can be input as additional premises to a target LM. We refer to this framework as DeduCE—Deductive Consistency Evaluation of LMs.

Importantly, the DeduCE framework requires a single valid solution for a problem, not the full ideal reasoner. Following ~\citep{gsm-symbolic}, we construct novel problems by altering variable values in the original benchmark problem. To explain LMs' decreased accuracy on novel problems, we provide a definition for deductive consistency that is parameterized by the number of input premises and output steps. Using this definition, we study how consistency changes as 1) partial CoTs include larger input premises, and 2) the task involves higher number of output steps to predict. We formalize both observations as the average slope or \textit{decay} in deductive consistency as number of input premises or output steps are increased.

We apply the proposed framework to evaluate multiple LMs on grade school math problems, using GSM8K\citep{gsm8k} and a synthetic dataset. As shown by past work, there is a significant drop in final accuracy on GSM8K between original and mutated problems. Our main findings are:
\begin{itemize}
    \item Deductive consistency of LMs is robust to the length of the input premises for GSM8K, but  sensitive to the number of reasoning hops. As the number of reasoning hops increases from 1 to 5, deductive consistency falls by 15-30\% (\autoref{fig:DC-hop_BaseDecay}) . Note that this effect was likely masked in original benchmark problems by memorization.
    \item Post-training methods like supervised fine-tuning and reinforcement learning enhance task-specific patterns rather than general deductive reasoning, even for the same underlying reasoning task.
    \item Other hypotheses, such as novel problems inducing a different language style in the solution or early errors that propagate, are unable to explain the difference in final accuracy between original and mutated problems. The maximum deviation seen is nearly 5\% (\autoref{fig:paraInt}) 
\end{itemize}

\input{related}

\input{methodology}

\input{evaluation}

\input{gsm8k}
\input{SynDeduct}

\input{Ablations}
\section{Conclusion}


We propose a metric for evaluating AI deductive reasoning using only text-based reasoning traces, enabling broad domain applicability.Our experiments reveal contrasting fine-tuning effects on synthetic (SynDeduct) and real-world (GSM8K) benchmarks, highlighting key trade-offs.

\section*{Impact Statement}
This paper presents work whose goal is to advance the reliability of AI reasoning. 
We expect that insights from our work can be used to improve reliability of AI reasoning, leading to a positive impact on downstream applications by avoiding reasoning bugs. 

\bibliography{iclr2025_conference}
\bibliographystyle{iclr2025_conference}

\newpage
\appendix

\input{Appendix}

\end{document}

%% file: math_commands.tex
\usepackage{amsmath,amsfonts,bm}

\def\eqref#1{equation~\ref{#1}}

\def\1{\bm{1}}

\DeclareMathAlphabet{\mathsfit}{\encodingdefault}{\sfdefault}{m}{sl}
\SetMathAlphabet{\mathsfit}{bold}{\encodingdefault}{\sfdefault}{bx}{n}



%% file: related.tex
\section{Related Work}

\textbf{Evaluating reasoning beyond memorization.}
To avoid confounding in experiments due to dataset memorization, novel datasets based on synthetic data have been proposed~\citep{zhu2023dyval}. To keep real world relevance, recent work propose perturbed or novel versions of existing datasets, for math~\citep{gsm-symbolic, zhang2024dargdynamicevaluationlarge}, analogical reasoning \citep{lewis2024using,lewis2024evaluating}, and many other diverse tasks~\citep{wu2024reasoningreciting}.

\textbf{Metrics for reasoning beyond final answer's accuracy.}
\citep{xu2024largelanguagemodelsreally} four different types of metrics based on answer correctness and explanation correctness. \citep{seals2024evaluating} test deductive reasoning based on logical questions. We aim to provide a general metric for any reasoning task. 
 Another stream of work checks language models' ability to detect errors in a solution~\citep{zeng2024mrgsm8kmetareasoningbenchmarklarge} and fix any detected errors~\citep{singh2024exposingachillesheelevaluating}.

\textbf{Deductive reasoning in LLMs.}
LogicBench evaluates various models on natural language problems over propositional, first order, and non-monotonic logic~\citep{logicbench}. Other examples include analyzing categorical syllogisms~\citep{categorical} and proving theorems in intuitionistic  propositional logic~\citep{proplogic}.

%% file: methodology.tex
\section{Defining Deductive Consistency}
Given a consistent proof system \( \mathcal{S} = \langle \mathcal{L}, \mathcal{R}\rangle \), where \( \mathcal{L} \) is the underlying logical language and \( \mathcal{R} \) is the set of inference rules, let \( \bm{P} \subset \mathcal{L} \) denote the set of premises and \( \text{Th}(\bm{P}) \subseteq \mathcal{L} \) the deductive closure of \( \bm{P} \) under \( \mathcal{R} \). We assume access to a dataset of problems $(\bm{P}_j, t_j)_{j=1}^M \sim \mathcal{D}$ where $P_j$ denotes the set of premises and $t_j$ the target predicate to be proved in each problem $j$. Total number of problems in the dataset being $M$.

For example, consider a system with the language $\mathcal{L}$ of statements of the form $X\rightarrow Y$ and transitivity as a single inference rule, $W \rightarrow X, X\rightarrow Y \Rightarrow W \rightarrow Y $. A sample set of premises may be ``$A \rightarrow B; B\rightarrow C; B \rightarrow D; C \rightarrow F; D \rightarrow F; E\rightarrow F; F \rightarrow G$'', and a target predicate to be proved be ``$A \rightarrow G$''. Assume that a reasoning system $\mathcal{A}$ (e.g., an AI reasoning model) produces the (incorrect) proof, $A \rightarrow D; A\rightarrow E; E \rightarrow G$ to conclude $A \rightarrow G$. Beyond final accuracy, to evaluate the reasoning system's steps \( \mathcal{A} \) on such problems, we define the Deductive Consistency metric.
\subsection{Deductive consistency given a complete proof system}
Consistency evaluates the extent to which a reasoning system \( \mathcal{A} \) agrees with the reference proof system $\mathcal{S}$. 
For each problem $d\sim \mathcal{D}$, where $d=(\bm{P},t)$, we generate a proof $\mathcal{A}(\bm{P},t)$ using $\mathcal{A}$. A simple way of measuring reasoning performance may be to compute per-predicate accuracy. For each \( X_i \in \mathcal{A}(\bm{P},t) \):
 $Cons(d_j) = \frac{\sum_i I_{X_i}}{|\mathcal{A}(\bm{P}_j,t_j)|}$
where $I_z$ is an indicator function, 1 whenever $z$ is correctly inferred by $\mathcal{A}$ and zero otherwise. For each $X_i$, we use the complete proof system to verify whether it is true or not, given $\bm{P}$ and $\bm{X}_k:\{k:1,2,..i-1\}$. For the example proof above, the metric will be 0.67 because the second predicate is incorrect.

A key part of deductive reasoning is to process multiple input premises and determine the next correct predicates. Therefore, we extend the above metric to include longer input premises than in an original problem. We do so by sampling a (correct) proof from the  reference proof system and adding the first $k$ steps of the proof to the input premises $\bm{P}$. The input premises now become $\bm{P}'=\bm{P} \cup \mathcal{S}_{k}(\bm{P},t)$ where $\mathcal{S}_{k}(\bm{P},t)$ is the first $k$ steps of the proof. 
Then, as k increases, we obtain a measure of how well a reasoning system can handle larger input premises. Let $X'_i$ be the proof steps generated by  $\mathcal{A}(\bm{P}\cup \mathcal{S}_k(\bm{P},t), t)$. 
The reasoning system's goal is to complete the proof. 
$$ Cons(d_j,k) = \frac{\sum_{i=k+1}^{N} I_{X'_{i}}}{|\mathcal{A}(\bm{P}'_j,t_j)|}$$

Continuing our example with $k=1$, the reference proof system may add the first step, ``$A \rightarrow C$'' and let the target system $\mathcal{A}$ complete the rest.  Here, the system may produce a  faulty proof as before, $A\rightarrow C; A \rightarrow E; A \rightarrow G$. However, with $k=2$ and adding the first two steps  ``$A \rightarrow C; A \rightarrow F$'', the system $\mathcal{A}$ may produce $A\rightarrow C; A \rightarrow F; A \rightarrow G$, which is a correct proof.   

However, the above metric has a right censoring issue~\citep{censor}: this measure of deductive consistency  depends trivially on the number of input premises.  In general, the difficulty of a proof is associated with the number of inference rules required to complete it. If many premises are already provided, the number of inference rules to reach the target predicate decreases and the problem becomes simpler.  As a result, if we see an increase in deductive consistency as the number of reference proof steps are increased (as we see for the example above), it may simply be due to the fewer steps that need to be predicted, rather than due to the reasoning system's improved consistency after access to the reference system's guidance for the first few steps.  Therefore, we also introduce a \textit{hops} parameter, denoting the number of inference rules (steps) until which we evaluate the reasoning system.
$$ DedCons(k, l) = \frac{ \sum_{d_j \sim \mathcal{D}} I_{X'_{k+l+1,j}}}{M}$$

Compared to final accuracy, a key benefit of the our formulation is that we obtain multiple premises and evaluation sets from a single problem instance. This allows us to test a diverse set of deductive tasks even from a small number of problem instances.

\subsection{Deductive Consistency given a reference proof}
While the above metric works for a \textit{complete} proof system as the reference, in practice it is more common to have access to a limited reference system that can only generate a single proof $\mathcal{S}(\bm{P}, t)$ given a problem. 
Therefore, we now assume access to a reference proof system that given a set of premises $\bm{P}$ and a target predicate $t$, can generate a proof involving predicates \( \mathcal{S}(\bm{P},t) \subseteq \text{Th}(\bm{P}) \), representing the predicates within the closure that were proved while proving the main result for $t$. We call such a proof as the \textit{reference solution}.
Continuing the transitivity example, it would mean that we only have access to a reference proof solution, $A \rightarrow C; A \rightarrow F; A \rightarrow G$, but cannot assess the validity of a predicate outside it such as $A \rightarrow D$.

Given a reasoning system's proof $\mathcal{A}(\bm{P}, t)$, this implies that we can only verify the predicates that are also present in $\mathcal{S}(\bm{P},t)$. We therefore orient the deductive consistency metric to focus on the fraction of the verifiable predicates that are proved by $\mathcal{A}$.  For each $Z_i \in \mathcal{S}(\bm{P}, t)$
$$ DedCons(k, l) = \frac{\sum_{d_j \sim \mathcal{D}} I_{Z'_{k+l+1} \in \mathcal{A}(\bm{P}'_j,t_j)}}{M}$$

where the numerator is an indicator function checking whether a given predicate $Z_i \in \mathcal{S}(\bm{P},t)$ is also included in the proof by $\mathcal{A}$. Note that the above metric introduces a bias because the reasoning system $\mathcal{A}$ may generate (true) predicates that are not in the reference solution (there can be multiple ways to solve the same problem). For instance, if a reasoning system produces a valid proof, $A \rightarrow D; A \rightarrow F; A \rightarrow G$, it will not have consistency=1 because the first step $A \rightarrow D$ is not a part of the reference solution $\mathcal{S}(\bm{P},t)$.

In such cases,  the consistency metric above can under-estimate the deductive consistency--the reasoning system may be penalized for a producing a valid solution because its steps are different than that of the reference system. Hence, we also introduce a metric for  coverage. 
The \textbf{Coverage} metric is defined as the expected proportion of variables in \( \mathcal{S}(\bm{P}, t) \) inferred by \( \mathcal{A} \). Let $\bm{V}_{\mathcal{S}(\bm{P}, t)}$ be the variables included in the reference solution. Then coverage is 


$$ Coverage =  \frac{\sum_{d_j \sim \mathcal{D}} | \bm{V}_{\mathcal{S}(\bm{P_j}, t)}  \cap \bm{V}_{\mathcal{A}(\bm{P_j}, t)}|} {\sum_{d_j \sim \mathcal{D}}|\bm{V}_{\mathcal{A}(\bm{P_j}, t)}|}$$

Thus, coverage measures how reliably verification of the set of predicates \( V \) measures consistency. When the coverage is high, \textbf{an ideal reasoning system's deductive consistency should be a constant close to 1, independent of the number of premises $k$ and the number of hops $l$}. To measure this we define decay w.r.t. hops as follows:

\[
\mu_\ell = \frac{1}{\ell_{\max}+1} \sum_{\ell=0}^{\ell_{\max}} \frac{\ell}{\ell_{\max}}, \quad 
\mu = \frac{1}{\ell_{\max}+1} \sum_{\ell=0}^{\ell_{\max}} \mathbb{E}_k \left[ \text{DedCons}(k, \ell) \right]
\]

\[
\text{Decay} = -\frac{\sum\limits_{\ell=0}^{\ell_{\max}} \left( \frac{\ell}{\ell_{\max}} - \mu_\ell \right) \left( \mathbb{E}_k \left[ \text{DedCons}(k, \ell) \right] - \mu \right)}{\sum\limits_{\ell=0}^{\ell_{\max}} \left( \frac{\ell}{\ell_{\max}} - \mu_\ell \right)^2}
\]

where $\ell_{\max} $ represents maximum number of hops in dataset.


%% file: evaluation.tex
\section{Evaluating Deductive Consistency for LLMs}
\label{Evaluating Deductive Consistency for LLMs}
As noted above, we need at least one reference solution for a reasoning problem to evaluate deductive consistency. Given a benchmark reasoning dataset, we now provide a method to obtain such solutions and evaluate deductive consistency. 

We use auxiliary expert LMs to help with transformation tasks. Specifically, we use \textbf{Code Generation LM} that generates executable reasoning graphs (Python Code), a \textbf{Templatization LM} that defines variable templates, and a \textbf{Variable Extraction LM} (Parser) that extracts predicate values for evaluation. The \textbf{Subject LM} is the model under evaluation. The entire pipeline is shown in Figure~\ref{fig:pipeline}.

\subsection{Generating correct solution for a benchmark problem}
While generating the correct solution for a reasoning problem is hard, an expert LM may be able to generate candidate solutions. In particular, its capability for generating good candidate solutions may be higher for the original benchmark problem. We therefore adopt a \textit{generate-then-verify} approach to select problems on which we can obtain a correct solution. We employ specific LMs to produce Code and templatized COT. The templatized COT is generated first, followed by the code. To build confidence in our candidate solution, we rely on the concept of internal consistency. We represent the solution in semantically equivalent forms, such as code and a templatetized version of the CoT, and perform multiple sanity checks to ensure its correctness.

One key sanity check involves verifying the equivalence between the code representation and the templatized Chain of Thought (tCoT). The variables in the code and the placeholders in the COT are equivalent. We validate this by ensuring that, given the same factual input, both representations yield identical variable values at each step. Once confirmed consistent with each other, the code serves as a symbolic representation of both the problem and its correct solution.

\subsection{Generating a novel problem by perturbing the benchmark problem}
Given the reasoning gap between existing benchmark problems and novel problems created by perturbation, we evaluate deductive consistency on novel problems only. This avoids any memorization concerns. In the current work, we adopt a simple perturbation: changing the values of variables in the problem statement. Other perturbations, such as changing variable names and adding irrelevant info~\citep{gsm-symbolic} can be easily added.

Using the template generated earlier, we sample new values for the variables in the template. These sampled values serve as inputs to the reasoning code, which computes the corresponding solution for the novel problem. We substitute the computed values back into the template Chain of Thought (tCoT) and template Question (tQ) to produce modified questions (Q') and their reasoning steps (COT'). COT' provides a detailed, step-by-step chain of reasoning for evaluation.

\subsection{Evaluating deductive consistency}
We evaluate deductive consistency by generating a perturbed question (Q') by modifying the seed premises and deriving its reasoning steps (COT) using the templatized Chain of Thought (tCoT). We then substitute intermediate premises into placeholders in tCoT by evaluating the Code with the same premises that generated the perturbed questions.

Through this process, we traverse a computation graph that serves as a proof Directed Acyclic Graph (DAG). Each link in the DAG represents a reasoning step, analogous to the transitivity of premises in a formal proof. For example, the sequence A → C; A → F; A → G encompasses the intermediate steps required to reach the target premise or the variable of interest.

We define a “hop” as the number of edges needed to progress from a specific premise to the target premise. Conversely, we define a “prefix” as the number of premises we provide to the model in advance ($k$). Using the template, we supply the LLM with $k$ steps worth of premises and then prompt it to solve the remaining steps—$l$ hops—needed to arrive at the final conclusion. By varying $k$ and $l$, we can systematically evaluate the model’s performance across different levels of partial information and reasoning depth.

\subsection{Reasoning evaluation on synthetic datasets}
\textbf{Why Use Synthetic Dataset?} Synthetic datasets offer a controlled framework for evaluating deductive accuracy, as all data points are generated according to predefined rules with precisely derived ground truth. This setup enables meticulous regulation of the underlying computation graph, allowing us to specify the total number of reasoning steps (i.e., edges in the graph) needed to arrive at the final answer. 

Furthermore, it becomes straightforward to craft questions that traverse designated nodes in a prescribed order. For example, a path can be orchestrated to move from the initial premise to an intermediate node A (“prefix k”) and then from node A to the target node B (“hop l”), such that Distance(seed premise, B) = k+l. This level of control over the graph’s structure proves highly valuable for generating datasets with specific properties and systematically assessing deductive performance.

\textbf{Dataset Generation}
In constructing of SynDeduct dataset, we begin by sampling a set of DAGs according to parameters that define constants, variable distributions, and arithmetic operators. We then extract paths from each DAG as programmatically computed reasoning traces. The resulting ground-truth derivations are converted into Chain of Thought representations by applying a set of verbalization templates, yielding readable textual explanations. Unlike in GSM8K, our approach does not require code generation or templated Chains of Thought, as the underlying computation graph is already available.

We quantify a path’s difficulty by counting the number of reasoning steps (graph edges) it takes to move from the initial (base) node to the final (target) node. Accordingly, we generate N sets of questions, where the n-th set contains questions that require n steps.

To accommodate varying input-premise lengths, we create additional questions by progressively appending segments of the ground-truth reasoning chain to converge on the same target premise. We then place these questions into bins based on how many hops are needed, intermixing different prefix lengths within each bin. This organization yields n bins, each focused on questions requiring n hops but differing in the prefixed portion of the chain. Such binning enables robust averaging of model performance for varying prefix lengths within the same number of steps. Details are present in Appendix \ref{subsec:synArt}.

%% file: gsm8k.tex
\section{Results: Math reasoning on GSM8K }
\subsection{Evaluation Setup}

\textbf{Dataset Statistics.} A subset of 1000 problems from GSM8K is randomly chosen. The responses of LMs under evaluation are filtered as described in \autoref{Evaluating Deductive Consistency for LLMs}. Problem instances common across the models are collected and used as final dataset that will be used to evaluate these models. This consists of 165 problem instances.

\textbf{LMs in DeduCE pipeline.} We use LLama-3-70B-Instruct LM as the templatizer, code generation and variable extractor. We find that LMs such as LLama-3-70B-Instruct are reasonably capable at templatization,  obtaining a failure rate (unable to generate json) close to 30\%, which we filter out. The additional sanity checks ensures that we have high quality dataset for evaluation. 

\textbf{Models under evaluation.} We evaluate the following LMs: Phi-3.5-mini-instruct, Phi-4, Qwen2.5-Math-7B-Instruct, Qwen2.5-Math-72B-Instruct, Llama-3.3-70B-Instruct, Llama-3-8B-Instruct. \textit{All models are Instruct tuned. Model suffixes will be truncated in plots}.


\begin{figure*}[t]
    \centering
    \begin{minipage}{\textwidth}  
        \centering
        {%
            \includegraphics[width=0.48\textwidth]{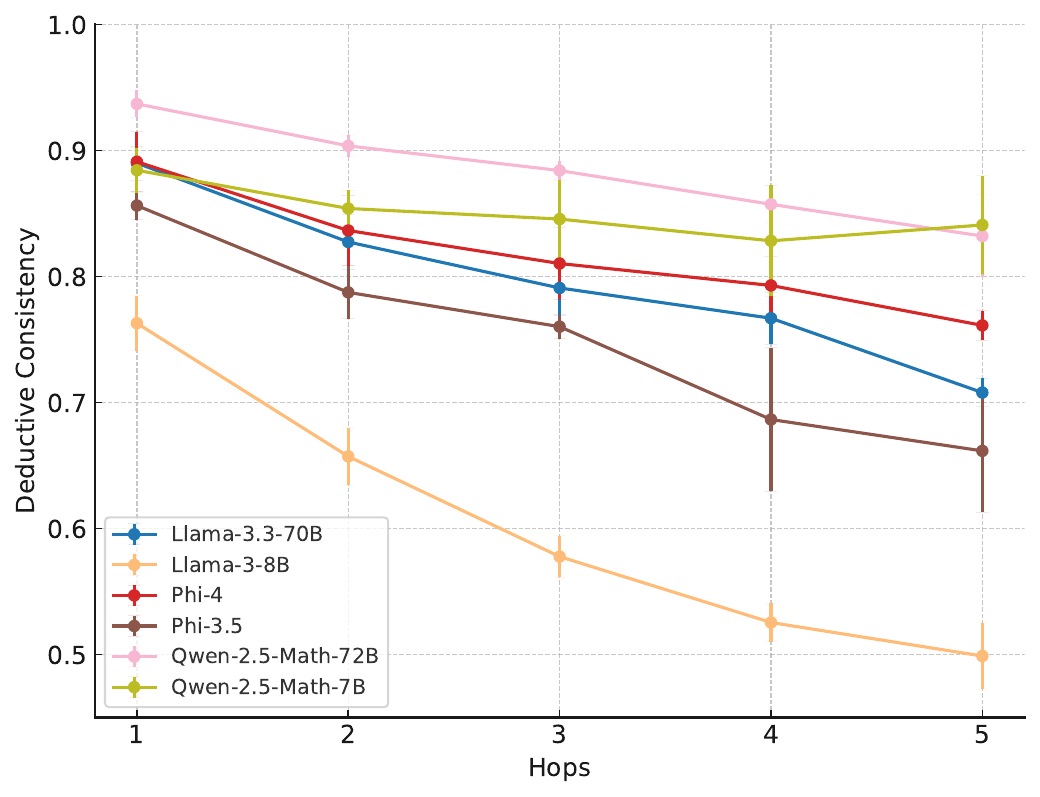}%
                \label{fig:DC_orig}
        }
       {%
            \includegraphics[width=0.48\textwidth]{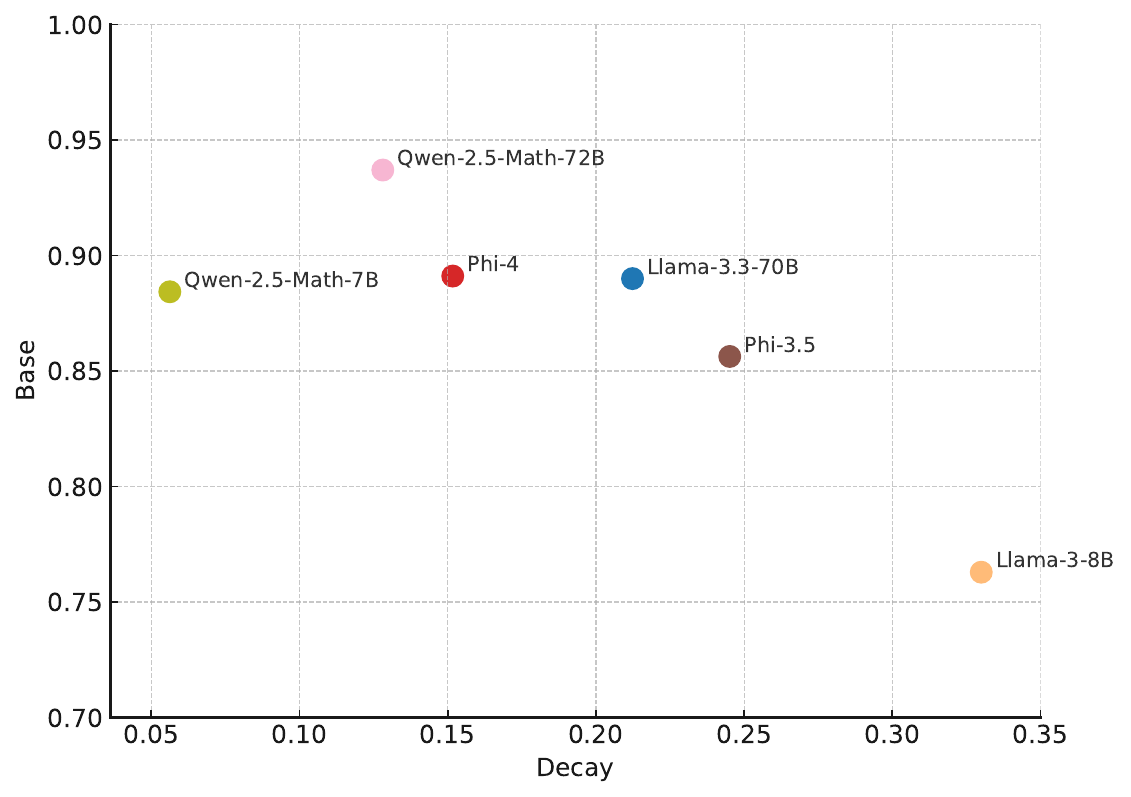}%
             \label{fig:base-decay-1}
        }
        
        \caption{\textbf{Left:} Deductive Consistency vs. Reasoning Hops across models. \textbf{Right:} Base deductive consistency vs. Decay.The premises are set using the Original paraphrase setting.}
        \label{fig:DC-hop_BaseDecay}
    \end{minipage}
\end{figure*}

\subsection{Hypotheses}
Based on the deductive consistency metric, we can formulate some hypotheses on why final accuracy decreases for novel math problems, as reported in past work~\citep{gsm-symbolic,srivastava2024functional}. We also test the hypothesis on the effects of language style on reasoning \citep{han2024surfacestructurecausalassessment}.

\textbf{H0}: Novel Problems induce error in chain of thought reasoning. \textbf{H1}: Novel problems induce early errors in the math computation, which propagate to lead to an incorrect solution. \textbf{H2}:  LMs have a significant decay in reasoning ability as the number of premises or hops increase. 
\textbf{H3}:  Novel problems induce a style change in the CoT answers, which may lead to faulty reasoning and hence incorrect final answer.

To decide between hypotheses, we create four kinds of premises that can be added. The first is the original (\textit{eg: Yasna has 60 + 12 = 72 pages to read.}), sourced from subject LMs answers on the original benchmark problem. 
Others are different paraphrases of the original style. They are explained in Appendix \ref{App:Paraphrase}

\subsection{Deductive reasoning decays with number of hops}

\textbf{Coverage} Table~\ref{tab:coverage} shows that the coverage is high across all premises. Given a LM, this implies that the intermediate variables inferred in the solution for the novel problem are almost the same as the variables inferred in the solution for the original benchmark problem. Therefore the code obtained from reasoning code generator is reliable as a reference proof. So we can go ahead with interpreting the consistency results.

\textbf{H0}: We find that the deductive consistency as a function of hops on the original benchmark achieves a constant value of 1 across all models. When deductive consistency is computed on perturbed problem a, we find it to be significantly lower (see \autoref{fig:DC-hop_BaseDecay}). This indicates \textbf{Memorization Effect} of the benchmark, validating \textbf{H0}.

\textbf{H1}:
Mean deductive consistency is computed by averaging predicate consistency across prefixes for a given hop.
We include only hops where the ratio of single-premise samples for the given hop to those with premise-length of 1 hop-1 exceeds 20\%, ensuring sufficient data for reliable estimates. Our findings reveal that consistency remains high for the first hop, contradicting hypothesis by demonstrating that models correctly answer the first step. While novel problems do not induce early errors, we observe frequent computational errors in model responses. These errors propagate, providing evidence for part of hypothesis. Detailed error analysis, is in Appendix \ref{App:gsm8kerror}.

\textbf{H2}: The second key result is \textit{decay} in deductive consistency as hops increase, which was masked due to memorization of the original benchmark.  We characterize this \textbf{decay} as negative of the slope of the best fit line in \autoref{fig:DC-hop_BaseDecay} with hops normalized between 0 and 1 and \textbf{base} refers to the deductive consistency of the reasoning model with 1 Hop. An ideal model must achieve zero decay and a base value of one.

We find that larger models (Qwen-Math-72B-Instruct, and Llama-3.3-70B-Instruct), models trained on synthetic data (Phi-4) as well as math-specific models (Qwen-2.5-Math-72B-Instruct, and Qwen-2.5-Math-72B-Instruct) do achieve greater base values. However, even these models show significant decay in the deductive consistency as the number of hops increases. Smaller models like Llama-3-8B-Instruct and Phi-3.5-mini-instruct perform poorly with lower base values and Llama-3-8B-Instruct exhibits a high decay value compared to other models (also see \autoref{fig:DC-hop_BaseDecay}). 

Importantly, deductive consistency does not vary much as the length of input premises are changed. We observe lower variance in mean deductive consistency as a function of prefix as seen in \autoref{fig:deductive_consistency_prefix} . So we have partial evidence of hypothesis: it depends on hops, but not on the premises. A caveat is that due to the simplicity of the GSM8K problems, the maximum premise length we could evaluate on is 7.

\textbf{H3}: We observe slight decrease in base values across models due to {impact of language style}. While it is expected that the original benchmark's style should have highest accuracy, the variation across paraphrases is not high. Even though, on performing t-test p-values values were significant (at 0.05 significance level), Cohens' effects sizes were too small to consider (\textless0.1). We find weak evidence for \textbf{H3} referring to \autoref{fig:Base_Decay_Combined} in Appendix.  


%% file: SynDeduct.tex
\section{Results: Evaluation on a  Synthetic dataset}
To validate the conclusions from GSM8K, we now evaluate deductive consistency on a synthetic dataset. In particular, the problems are designed such that the solutions involve a large number of hops spread across prefixes. \textit{All models are Instruct tuned}.


\subsection{Results on SynDeduct}

\begin{figure}[h]
    
    \centering
    \includegraphics[width=0.7\textwidth]{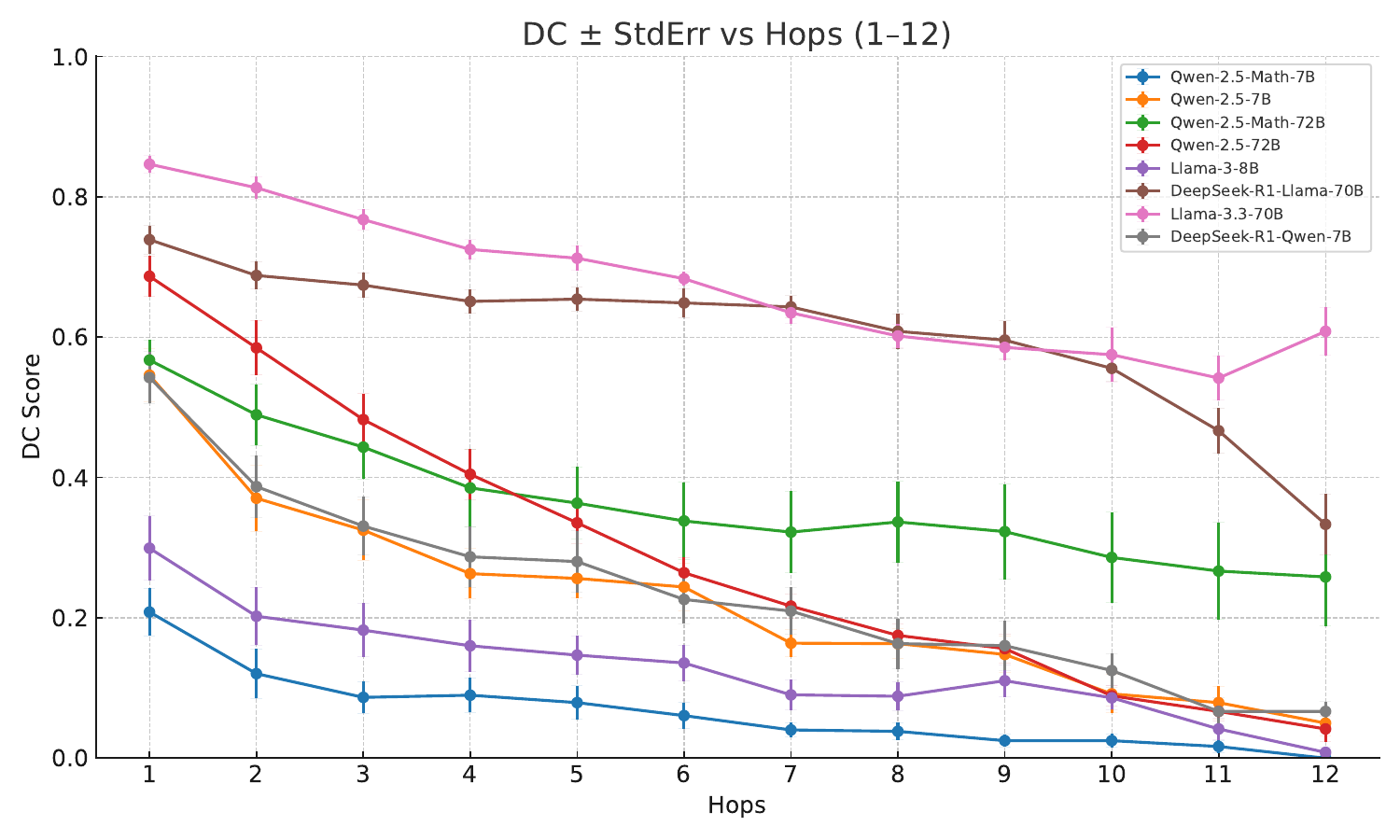}
    \caption{Accuracy v/s Hops. Each Hop bin has roughly same variation of Prefixes. Y-Ticks on each data-point is Standard Error for Accuracy. For Coverage refer to Appendix \autoref{fig:coverage_syndeduct}.}
    \label{fig:SynDeduct}
\end{figure}


The general trend of decreasing deductive consistency over hops supports our results in GSM8K. Our findings in ~\autoref{fig:SynDeduct} suggests larger models demonstrate greater resilience to increases in the number of hops, while smaller models—such as Llama-3-8B-Instruct—experience a substantial drop in performance.
One contributing factor may be the larger maximum token limit, which allows these models to accommodate more extensive reasoning chains. However, for queries with increasingly longer chains, the input size expands significantly, risking the approach of token length limits and thereby degrading performance. This trend becomes evident when examining accuracy versus prefix length across multiple hops: as prefix segments grow, the overall task accuracy declines. We observe that larger instruction-tuned and SFT-tuned models exhibit lower decay with respect to prefix length. While Qwen-2.5-Math-72B shows improved deductive consistency at shorter prefixes, these gains diminish rapidly with increasing prefix length—eventually falling below the performance of its base model.(Appendix \autoref{tab:prefix_dc_vs_hops_1_6}, \autoref{tab:prefix_dc_vs_hops_7_12})

%% file: Ablations.tex
\section{Ablations}
\subsection{Distillation and RL Tuned Models}
To better understand the impact of different fine-tuning strategies on deductive reasoning, 
we focus on two primary fine-tuning approaches: iterative fine-tuning with reinforcement learning (RL-training) and supervised fine-tuning (SFT). 

For RL-based and iterative fine-tuning models, we analyze Qwen-2.5-Math-Instruct in both its 7B and 72B variants \citep{qwenAtg}, comparing them against their respective base models. Similarly, for SFT-based tuning, we compare R1 distilled models to their base counterparts \citep{deepseek}, namely DeepSeek-R1-Distill-Llama-70B and DeepSeek-R1-Distill-Qwen-7B. 

 RL fine tuning is more effective (\autoref{tab:first_syn}, \autoref{tab:first}) in reducing the decay of deductive consistency. RL fine-tuning shows minimal change in base deductive consistency on in-distribution datasets, as a drastic reduction on unseen datasets such as SynDeduct. SFT after training causes a decrease in deductive consistency and worsens decay for both data sets (\autoref{tab:second_syn}, \autoref{tab:second}). More work is required to study the extent of generalization that such post-training methods provide. In general, these findings emphasize that neither of the two post-training techniques is successful in improving deductive consistency between models and datasets.

\begin{table}[ht]
  \centering

  \begin{minipage}{.47\textwidth}
    \centering
    \resizebox{\textwidth}{!}{
    \begin{tabular}{ccc}
    \toprule
    \textbf{Model} & \textbf{Base} & \textbf{Decay} \\
    \midrule
    Qwen-2.5-72B-Instruct & 0.6868 & 0.0602  \\
    Qwen-2.5-Math-72B-Instruct & 0.5674 & 0.0273 \\
    \midrule
    Qwen-2.5-7B-Instruct & 0.5458 & 0.0432  \\
    Qwen-2.5-Math-7B-Instruct & 0.2083 & 0.0211  \\
    \bottomrule
    \end{tabular}
    }
    \caption{Ablation for RL post training on SynDeduct.}
    \label{tab:first_syn}
  \end{minipage}
  \hfill
  \begin{minipage}{.47\textwidth}
    \centering
    \resizebox{\textwidth}{!}{
    \begin{tabular}{ccc}
    \toprule
    \textbf{Model} & \textbf{Base} & \textbf{Decay}  \\
    \midrule
    Qwen-2.5-Math-7B-Instruct & 0.2083 & 0.0211  \\
    DeepSeek-R1-Distill-Qwen-7B & 0.5424 & 0.0381  \\
    \midrule
    Llama-3.3-70B-Instruct & 0.8465 & 0.0212  \\
    DeepSeek-R1-Distill-Llama-70B & 0.7389 & 0.0314  \\
    \bottomrule
    \end{tabular}
    }
    \caption{Ablation for SFT post training on SynDeduct.}
    \label{tab:second_syn}
  \end{minipage}
\end{table}

\begin{table}[ht]
  \centering

  \begin{minipage}{.47\textwidth}
    \centering
    \resizebox{\textwidth}{!}{
    \begin{tabular}{ccc}
    \toprule
    \textbf{Model} & \textbf{Base} & \textbf{Decay} \\
    \midrule
    Qwen-2.5-72B-Instruct & 0.9149 & 0.2339 \\
    Qwen-2.5-Math-72B-Instruct & 0.9164 & 0.1725 \\
    \midrule
    Qwen-2.5-7B-Instruct & 0.8881 & 0.1618 \\
    Qwen-2.5-Math-7B-Instruct & 0.8427 & 0.1189 \\
    \bottomrule
    \end{tabular}
    }
    \caption{Ablation for RL post training on GSM8k.}
    \label{tab:first}

  \end{minipage}
  \hfill
  \begin{minipage}{.47\textwidth}
    \centering
    \resizebox{\textwidth}{!}{
    \begin{tabular}{ccc}
    \toprule
    \textbf{Model} & \textbf{Base} & \textbf{Decay} \\
    \midrule
    Qwen-2.5-Math-7B-Instruct & 0.8509 & 0.0613 \\
    DeepSeek-R1-Distill-Qwen-7B & 0.8468 & 0.1957 \\
    \midrule
    Llama-3.3-70B-Instruct & 0.8532 & 0.20065 \\
    DeepSeek-R1-Distill-Llama-70B & 0.8366 & 0.24895 \\
    \bottomrule
    \end{tabular}}
    \caption{Ablation for SFT post training on GSM8k.}
    \label{tab:second}

  \end{minipage}
\end{table}

\subsection{Error Analysis}

\textbf{Evaluation method.} Model responses are grouped into five groups based on final accuracy on the mutated dataset. Accuracy here is computed over the set of mutated problems for each problem in the original benchmark. The groups are;
     \textit{ Group-1 : Accuracy $=$ 1 ;
     Group-2 : 1 $<$ Accuracy $\le$ 0.7 ;
     Group-3 : 0.7 $<$ Accuracy $\le$ 0.4 ;
     Group-4 : 0.4 $<$ Accuracy $<$ 0 ;
     Group-5 : Accuracy $=$ 0 }

\textbf{Error Categories.} We use GPT-4o as an evaluator. Calculation errors like arithmetic mistakes, as well as errors in rounding, along with error propagation 
Logic errors are wrong application of logic/rule/formula. Understanding errors are wrong assumption or contradiction of a given fact. These errors are seen in cases where the problem mentions scenarios that are far from real world such as there being 97 days in a week. In \autoref{tab:error_metrics} we report the frequency of error normalized by number of error responses in that group.

\textbf{Observation and Findings.} A higher proportion of calculation errors is observed relative to logical and comprehension errors. These calculation errors predominantly emerge during arithmetic operations within the chain-of-thought, and they propagate through subsequent reasoning steps.

Furthermore, models exhibit (pre-training) bias. They reproduce the original reasoning graph from the vanilla solution. This shows weak robustness from changes in reasoning structure. Logical errors stem from ambiguous natural language. For instance, the sentence “My brother is twice more older than me” should ideally be represented as: $myBrotherAge = myAge + myAge * 2.$ However, models typically interpret it as: $myBrotherAge = 2 * myAge$ which correctly corresponds to the unambiguous phrasing “My brother is twice as old as me.”. Refer to Appendix \ref{App:gsm8kerror} for details.

\begin{table}[h!]
    \centering
    \renewcommand{\arraystretch}{1.2}
    \resizebox{\textwidth}{!}{
    \begin{tabular}{lccc|ccc|ccc|ccc|ccc}
        \toprule
        \multirow{2}{*}{\textbf{Models}} & \multicolumn{3}{c|}{\textbf{Group-1}} & \multicolumn{3}{c|}{\textbf{Group-2}} & \multicolumn{3}{c|}{\textbf{Group-3}} & \multicolumn{3}{c|}{\textbf{Group-4}} & \multicolumn{3}{c}{\textbf{Group-5}} \\
        & Logical & Understanding & Calculation & Logical & Understanding & Calculation & Logical & Understanding & Calculation & Logical & Understanding & Calculation & Logical & Understanding & Calculation \\
        \midrule
        \textbf{Llama-3.3-70B} & 0.141 & 0.413 & 0.457 & 0.300 & 0.667 & 0.633 & 0.182 & 0.601 & 0.790 & 0.287 & 0.943 & 0.780 & 0.356 & 0.578 & 0.856 \\
        \textbf{Llama-3-8B} & 0.068 & 0.136 & 0.614 & 0.140 & 0.500 & 0.840 & 0.198 & 0.548 & 0.853 & 0.279 & 0.624 & 0.886 & 0.344 & 0.672 & 0.822 \\
        \textbf{Phi-4} & 0.036 & 0.175 & 0.130 & 0.128 & 0.368 & 0.248 & 0.229 & 0.702 & 0.550 & 0.261 & 0.513 & 0.704 & 0.369 & 0.946 & 0.754 \\
        \textbf{Phi-3.5} & 0.169 & 0.312 & 0.429 & 0.158 & 0.554 & 0.576 & 0.152 & 0.488 & 0.784 & 0.227 & 0.553 & 0.827 & 0.429 & 0.659 & 0.865 \\
        \textbf{Qwen-2.5-Math-72B} & 0.052 & 0.188 & 0.127 & 0.121 & 0.423 & 0.340 & 0.245 & 0.669 & 0.619 & 0.365 & 0.794 & 0.518 & 0.300 & 0.583 & 0.883 \\
        \textbf{Qwen-2.5-Math-7B} & 0.125 & 0.368 & 0.429 & 0.091 & 0.400 & 0.551 & 0.263 & 0.563 & 0.721 & 0.304 & 0.562 & 0.788 & 0.255 & 0.391 & 0.818 \\
        \bottomrule
    \end{tabular}}
    \caption{Error metrics for different models across dataset groups. Each cell reports Logical, Understanding, and Calculation errors separately.}
    \label{tab:error_metrics}
\end{table}

%% file: Appendix.tex
\section{Appendix}

\subsection{Details for GSM8K pipeline}
\textbf{Inference on original dataset} We sample a subset of GSM8K of size 1000. We prompt the LM under investigation to solve the question using the prompt template provided in \autoref{App:PromptsGSM8K}.

\textbf{Templatization and Code Generation} We templatize the question and LM CoT response using Llama-3-70B as Template Builder Agent. The model is prompted (as shown in \autoref{App:PromptsGSM8K}) to generate templatized question, templatized CoT answer (as well as chunk it into steps), explanation of variables of templates along with assignment of variables in question.

\textbf{Sanity Checks} We check that the code produced is an executable code, if the format of template generate is consistent with our reference template format, if the all variables in factual\_assignment are present in code. The generated code is executed with factual assignment as inputs for variables in question template and the value of other variables in code are checked to be consistent with the factual assignment in template. Further we have check if the final answer in response matches the ground truth answer in original dataset. If any of these checks fail then we remove that question from pipeline. For each model we now have a reduces set of questions that has passed sanity checks. We take intersection of such questions over multiple models to get a dataset on which we can evaluate all the models under consideration. This support set depends on the set of models being used in the experiments. 

\textbf{Mutation Details}We create mutated dataset by sampling the values of variables in question and executing the code with these newly sampled values to obtain assignment corresponding to other variables. Parameters for the sampler are (min-value,max-value,max-iter). If the factual assignment of a variable is integer, we sample from integers in the range (min-value,max-value), if factual assignment of a variable is decimal between 0 and 1, we uniformly sample from this range, else if it is any other decimal we sample a float from (min-valu,max-value). We try to make sure that all the variable assignments after positive. If not we rerun until we get a all positive assignment or we reach maximum iterations of the sampler. We substitute these values into template question and template CoT answer.We sample 10 mutated questions per question in original dataset. We create dataset with mutated question and varying length of mutated CoT answer present in LMs context. The number of steps from mutated template CoT answer is defined as prefix length. We collect the sampled variable assignments,mutated Question and Prefix into the mutated dataset.

\textbf{Inference on mutated dataset}We run inference of LMs on this mutated dataset. Since all LMs we evaluate are Instruction tuned, we use chat template. Mutated question is passed as user-content where as prefix is passed as assistant-content. We remove the \textless $\|$eot$\|$ \textgreater  token and let the generation continue as if the model were completing the generation.  

\textbf{Computation of Deductive consistency} The response of the model to mutated dataset is passed into a variable extraction LM which extracts value if the variables under consideration (ones in template) if present in response(see \autoref{App:PromptsGSM8K}). Hops are decided by the relative positions of variables under consideration in the template CoT. We check if these extracted values are consistent with code-generated values. We collect this data for every variable across mutated question and then group it by prefix and hop. We filter instances where the responses gives From this data structure we derive Deductive Consistency as function of hops or prefixes (as required).

\subsection{Para Phrasing Styles}
\label{App:Paraphrase}

\begin{itemize}
    \item Vanilla (Para-van): In this approach, the text is rephrased using conventional linguistic variations without incorporating any specialized semantic constraints. For example: \textit{Yasna's task involves 60 pages plus an additional 12 pages, which totals to 72 pages}
    \item Axiomatic (Para-ax): This method reformulates the statement as a set of axioms that articulate the underlying numerical relationships. An example is  : \textit{Axiom-1 (Addition): Given two numerical values x and y, the operation ADD(x, y) yields their arithmetic sum, thus ADD(60, 12) yields 72, which represents the total number of pages Yasna has to read.}
    \item Reverse (Para-rev): the sentence is restructured by inverting the typical cause-effect relationship—presenting the effect before providing the explanation for its cause. For instance, in the example: \textit{Yasna has to read the 72 pages, which is the sum of 60 and 12.} 
\end{itemize}

\subsection{Results - GSM8K }
Here we report the full results presented in the main paper. 

\begin{figure*}[t]
    \centering
    \begin{minipage}{\textwidth}  
        \centering
        {%
            \includegraphics[width=0.32\textwidth]{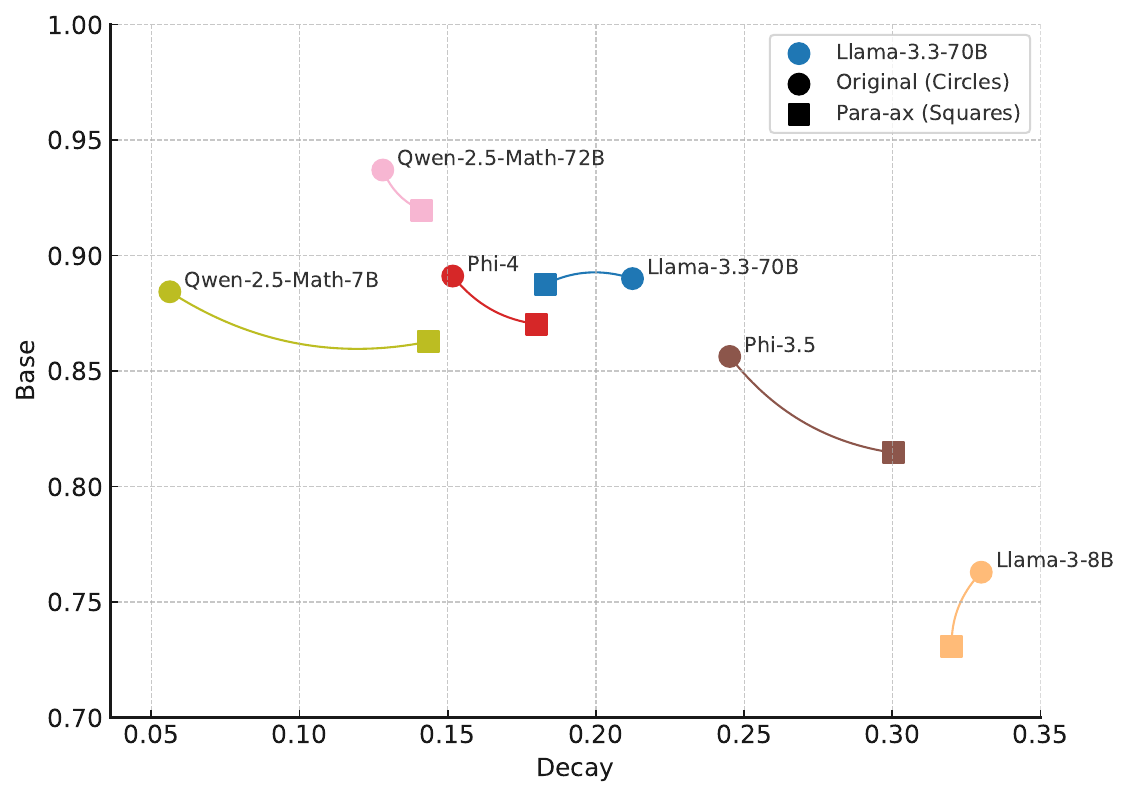}%
            \label{fig:Base_Decay_ParaAx}%
        }
       {%
            \includegraphics[width=0.32\textwidth]{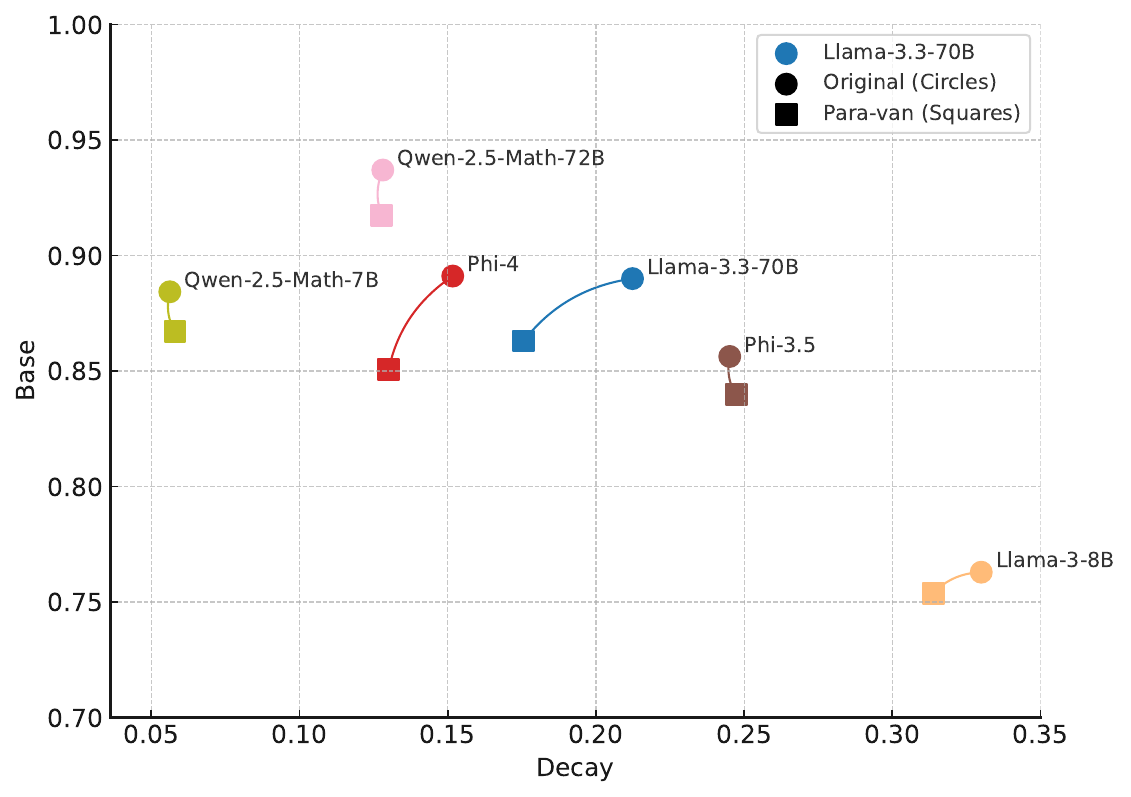}%
            \label{fig:Base_Decay_ParaVan}%
        }
        {%
            \includegraphics[width=0.32\textwidth]{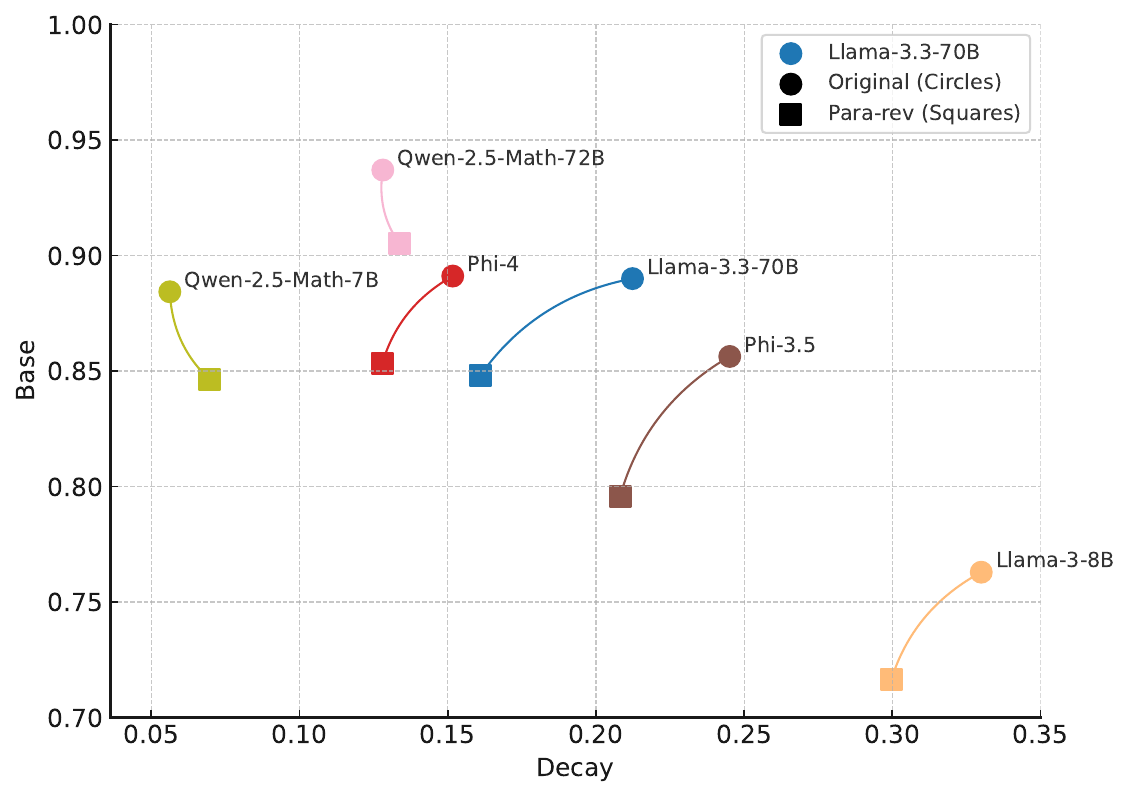}%
            \label{fig:Base_Decay_ParaRev}%
        }
        
        \caption{Comparison of Base vs. Decay Across Different Styles. Model name suffixes have been dropped for readability.}
        \label{fig:Base_Decay_Combined}
    \end{minipage}
\end{figure*}

\begin{table}[h]
    \centering
    \caption{Model Performance and Deductive Consistency Across Hops GSM8K}
    \resizebox{\textwidth}{!}{
    \begin{tabular}{lccccc}
        \toprule
        Models & Hop-1 & Hop-2 & Hop-3 & Hop-4 & Hop-5 \\
        \midrule
        Llama-3.3-70B & $0.89 \pm 0.0135$ & $0.8274 \pm 0.0219$ & $0.7909 \pm 0.0215$ & $0.7669 \pm 0.0206$ & $0.7079 \pm 0.0114$ \\
        Llama-3-8B & $0.7629 \pm 0.0217$ & $0.6572 \pm 0.0227$ & $0.5777 \pm 0.0165$ & $0.5254 \pm 0.0154$ & $0.4988 \pm 0.0264$ \\
        Phi-4 & $0.8911 \pm 0.0238$ & $0.8365 \pm 0.0278$ & $0.8103 \pm 0.0285$ & $0.7929 \pm 0.0228$ & $0.7612 \pm 0.0117$ \\
        Phi-3.5 & $0.8563 \pm 0.0114$ & $0.7874 \pm 0.0210$ & $0.7602 \pm 0.0096$ & $0.6865 \pm 0.0571$ & $0.6616 \pm 0.0488$ \\
        Qwen-2.5-Math-72B & $0.937 \pm 0.0108$ & $0.9037 \pm 0.0085$ & $0.8841 \pm 0.0073$ & $0.8573 \pm 0.0148$ & $0.8321 \pm 0.0328$ \\
        Qwen-2.5-Math-7B & $0.8843 \pm 0.0179$ & $0.854 \pm 0.0144$ & $0.8456 \pm 0.0307$ & $0.8283 \pm 0.044$ & $0.8409 \pm 0.039$ \\
        \bottomrule
    \end{tabular}}
\end{table}

\begin{table}[t]
    \centering
    \caption{Coverage across different language styles for the premises.}
    \resizebox{0.5\textwidth}{!}
    {
    \begin{tabular}{lcccc}
        \toprule
        Models & Original & Para-ax & Para-van & Para-rev \\
        \midrule
        Llama-3.3-70B & 0.9725 & 0.9639 & 0.9616 & 0.966 \\
        Llama-3-8B & 0.9669 & 0.9431 & 0.9543 & 0.9486 \\
        Phi-4 & 0.9849 & 0.9694 & 0.9759 & 0.9747 \\
        Phi-3.5 & 0.9684 & 0.9523 & 0.9649 & 0.9624 \\
        Qwen-2.5-Math-72B & 0.9888 & 0.9745 & 0.9862 & 0.985 \\
        Qwen-2.5-Math-7B & 0.9701 & 0.9442 & 0.9656 & 0.9648 \\
        \bottomrule
    \end{tabular}
    }
    \label{tab:coverage}
\end{table}
\begin{table}[h]
    \centering
    \caption{Para-ax: Model Performance and Deductive Consistency Across Hops}
    \resizebox{\textwidth}{!}{
    \begin{tabular}{lccccc}
        \toprule
        Models & Hop-1 & Hop-2 & Hop-3 & Hop-4 & Hop-5 \\
        \midrule
        Llama-3.3-70B & $0.8875 \pm 0.0158$ & $0.8083 \pm 0.0271$ & $0.7653 \pm 0.0282$ & $0.752 \pm 0.0189$ & $0.7328 \pm 0.004$ \\
        Llama-3-8B & $0.7309 \pm 0.0244$ & $0.6177 \pm 0.034$ & $0.5096 \pm 0.0305$ & $0.4799 \pm 0.0328$ & $0.4798 \pm 0.0173$ \\
        Phi-4 & $0.8703 \pm 0.0178$ & $0.8245 \pm 0.0177$ & $0.8055 \pm 0.0172$ & $0.7525 \pm 0.0275$ & $0.7263 \pm 0.0164$ \\
        Phi-3.5 & $0.8146 \pm 0.027$ & $0.694 \pm 0.0279$ & $0.6106 \pm 0.0426$ & $0.5795 \pm 0.0658$ & $0.5715 \pm 0.0424$ \\
        Qwen-2.5-Math-72B & $0.9196 \pm 0.0143$ & $0.851 \pm 0.0168$ & $0.8149 \pm 0.0194$ & $0.8038 \pm 0.0154$ & $0.8019 \pm 0.026$ \\
        Qwen-2.5-Math-7B & $0.8627 \pm 0.0192$ & $0.8001 \pm 0.0233$ & $0.7672 \pm 0.0216$ & $0.7418 \pm 0.0165$ & $0.7483 \pm 0.0148$ \\
        \bottomrule
    \end{tabular}}
\end{table}

\begin{table}[h]
    \centering
    \caption{Para-van: Model Performance and Deductive Consistency Across Hops}
    \resizebox{\textwidth}{!}{
    \begin{tabular}{lccccc}
        \toprule
        Models & Hop-1 & Hop-2 & Hop-3 & Hop-4 & Hop-5 \\
        \midrule
        Llama-3.3-70B & $0.863 \pm 0.0276$ & $0.79 \pm 0.0318$ & $0.7561 \pm 0.0277$ & $0.7242 \pm 0.0323$ & $0.7203 \pm 0.0103$ \\
        Llama-3-8B & $0.7538 \pm 0.0212$ & $0.612 \pm 0.0352$ & $0.5503 \pm 0.0207$ & $0.4997 \pm 0.0348$ & $0.4961 \pm 0.0275$ \\
        Phi-4 & $0.8505 \pm 0.0248$ & $0.8062 \pm 0.0271$ & $0.7856 \pm 0.0273$ & $0.7655 \pm 0.0356$ & $0.7408 \pm 0.0153$ \\
        Phi-3.5 & $0.8397 \pm 0.0189$ & $0.739 \pm 0.0203$ & $0.7178 \pm 0.0132$ & $0.6679 \pm 0.0366$ & $0.6279 \pm 0.0355$ \\
        Qwen-2.5-Math-72B & $0.9175 \pm 0.0144$ & $0.8758 \pm 0.0143$ & $0.8569 \pm 0.0138$ & $0.8427 \pm 0.0182$ & $0.8063 \pm 0.0284$ \\
        Qwen-2.5-Math-7B & $0.8671 \pm 0.0207$ & $0.8206 \pm 0.0192$ & $0.7841 \pm 0.0271$ & $0.7854 \pm 0.0246$ & $0.8267 \pm 0.0299$ \\
        \bottomrule
    \end{tabular}}
\end{table}

\begin{table}[h]
    \centering
    \caption{Para-rev: Model Performance and Deductive Consistency Across Hops}
    \resizebox{\textwidth}{!}{
    \begin{tabular}{lccccc}
        \toprule
        Models & Hop-1 & Hop-2 & Hop-3 & Hop-4 & Hop-5 \\
        \midrule
        Llama-3.3-70B & $0.8481 \pm 0.0222$ & $0.7828 \pm 0.0306$ & $0.7508 \pm 0.0313$ & $0.7148 \pm 0.0334$ & $0.7211 \pm 0.0097$ \\
        Llama-3-8B & $0.7163 \pm 0.0288$ & $0.5509 \pm 0.0504$ & $0.4612 \pm 0.0591$ & $0.4609 \pm 0.0371$ & $0.4617 \pm 0.025$ \\
        Phi-4 & $0.8532 \pm 0.0282$ & $0.8102 \pm 0.0289$ & $0.7874 \pm 0.0326$ & $0.7724 \pm 0.025$ & $0.7442 \pm 0.0122$ \\
        Phi-3.5 & $0.7958 \pm 0.0241$ & $0.7308 \pm 0.024$ & $0.6917 \pm 0.0165$ & $0.6351 \pm 0.0483$ & $0.6355 \pm 0.0431$ \\
        Qwen-2.5-Math-72B & $0.9051 \pm 0.017$ & $0.8636 \pm 0.0166$ & $0.8279 \pm 0.0172$ & $0.8105 \pm 0.0176$ & $0.7979 \pm 0.0258$ \\
        Qwen-2.5-Math-7B & $0.8463 \pm 0.0219$ & $0.7914 \pm 0.0227$ & $0.7298 \pm 0.0364$ & $0.7543 \pm 0.0124$ & $0.7952 \pm 0.0201$ \\
        \bottomrule
    \end{tabular}}
\end{table}

\begin{table}[h]
    \centering
    \caption{Decay and Base Values Across Interventions}
    \resizebox{\textwidth}{!}{
    \begin{tabular}{lcccccccc}
        \toprule
        Models & \multicolumn{2}{c}{Original} & \multicolumn{2}{c}{Para-ax} & \multicolumn{2}{c}{Para-van} & \multicolumn{2}{c}{Para-rev} \\
        \cmidrule(r){2-3} \cmidrule(r){4-5} \cmidrule(r){6-7} \cmidrule(r){8-9}
         & Decay & Base & Decay & Base & Decay & Base & Decay & Base \\
        \midrule
        Llama-3.3-70B & 0.21235 & 0.89 & 0.18285 & 0.8875 & 0.1756 & 0.863 & 0.161 & 0.8481 \\
        Llama-3-8B & 0.33 & 0.7629 & 0.32 & 0.7309 & 0.31385 & 0.7538 & 0.2996 & 0.7163 \\
        Phi-4 & 0.1517 & 0.8911 & 0.18 & 0.8703 & 0.13005 & 0.8505 & 0.1279 & 0.8532 \\
        Phi-3.5 & 0.24515 & 0.8563 & 0.30035 & 0.8146 & 0.24735 & 0.8397 & 0.20815 & 0.7958 \\
        Qwen-2.5-Math-72B & 0.1281 & 0.937 & 0.1413 & 0.9196 & 0.12775 & 0.9175 & 0.13375 & 0.9051 \\
        Qwen-2.5-Math-7B & 0.05625 & 0.8843 & 0.14355 & 0.8627 & 0.058 & 0.8671 & 0.06965 & 0.8463 \\
        \bottomrule
    \end{tabular}}
\end{table}

\begin{table}[h]
    \centering
    \caption{Deductive Consistency ± Standard Error vs. Prefix Length for Different Models}
    \resizebox{\textwidth}{!}{
    \begin{tabular}{lccccc}
        \toprule
        \textbf{Model} & \textbf{Prefix 1} & \textbf{Prefix 2} & \textbf{Prefix 3} & \textbf{Prefix 4} & \textbf{Prefix 5} \\
        \midrule
        Phi-3.5 & $0.7664 \pm 0.0304$ & $0.7772 \pm 0.0255$ & $0.7767 \pm 0.0224$ & $0.8153 \pm 0.0247$ & $0.7707 \pm 0.0215$ \\
        Qwen-2.5-Math-7B & $0.8549 \pm 0.0268$ & $0.8366 \pm 0.0195$ & $0.8510 \pm 0.0243$ & $0.8641 \pm 0.0329$ & $0.8922 \pm 0.0383$ \\
        Qwen-2.5-Math & $0.8802 \pm 0.0200$ & $0.8704 \pm 0.0223$ & $0.8803 \pm 0.0166$ & $0.9067 \pm 0.0117$ & $0.8967 \pm 0.0161$ \\
        Llama-3-8B & $0.5884 \pm 0.0580$ & $0.5826 \pm 0.0523$ & $0.5932 \pm 0.0466$ & $0.5984 \pm 0.0306$ & $0.6637 \pm 0.0330$ \\
        Llama-3.3-70B & $0.7981 \pm 0.0342$ & $0.7929 \pm 0.0322$ & $0.7937 \pm 0.0369$ & $0.7952 \pm 0.0344$ & $0.7905 \pm 0.0322$ \\
        Phi-4 & $0.8254 \pm 0.0303$ & $0.8277 \pm 0.0274$ & $0.8309 \pm 0.0311$ & $0.8199 \pm 0.0440$ & $0.7996 \pm 0.0444$ \\
        \bottomrule
    \end{tabular}}
    \label{tab:deductive_consistency_stderr}
\end{table}
\clearpage

\begin{figure}[h]
\centering

\begin{minipage}[t]{\textwidth}
\captionsetup{type=table}
\captionof{table}{Ablation 1: Deductive Consistency vs. Hops}
\resizebox{\linewidth}{!}{
\begin{tabular}{lccccc}
\toprule
Hops & 1 & 2 & 3 & 4 & 5 \\
\midrule
Qwen-2.5-72B & $0.9149 \pm 0.0123$ & $0.861 \pm 0.022$ & $0.8078 \pm 0.011$ & $0.7656 \pm 0.0203$ & $0.7287 \pm 0.02$ \\
Qwen-2.5-Math-72B & $0.9164 \pm 0.0091$ & $0.8739 \pm 0.0138$ & $0.8305 \pm 0.0272$ & $0.7895 \pm 0.0402$ & $0.7861 \pm 0.0584$ \\
\bottomrule
\end{tabular}}
\end{minipage}
\hfill
\begin{minipage}[t]{\textwidth}
\captionsetup{type=table}
\captionof{table}{Ablation 2: Deductive Consistency vs. Hops}
\resizebox{\linewidth}{!}{
\begin{tabular}{lccccc}
\toprule
Model & Hop 1 & Hop 2 & Hop 3 & Hop 4 & Hop 5 \\
\midrule
Qwen-2.5-7B & $0.8881 \pm 0.0176$ & $0.8453 \pm 0.0246$ & $0.8101 \pm 0.0222$ & $0.7738 \pm 0.0281$ & $0.7620 \pm 0.0352$ \\
Qwen-2.5-Math-7B & $0.8427 \pm 0.0156$ & $0.8021 \pm 0.0131$ & $0.7739 \pm 0.0142$ & $0.7499 \pm 0.0251$ & $0.7499 \pm 0.0306$ \\
\bottomrule
\end{tabular}}
\end{minipage}

\vspace{1em}

\begin{minipage}[t]{\textwidth}
\captionsetup{type=table}
\captionof{table}{Ablation 3: Deductive Consistency vs. Hops}
\resizebox{\linewidth}{!}{
\begin{tabular}{lccccc}
\toprule
Hops & 1 & 2 & 3 & 4 & 5 \\
\midrule
Qwen-2.5-Math-7B & $0.8509 \pm 0.018$ & $0.8093 \pm 0.0168$ & $0.8002 \pm 0.0198$ & $0.7957 \pm 0.0382$ & $0.7964 \pm 0.0458$ \\
DeepSeek-R1-Distill-Qwen-7B & $0.8468 \pm 0.0271$ & $0.7989 \pm 0.0391$ & $0.7451 \pm 0.0414$ & $0.7309 \pm 0.0566$ & $0.6851 \pm 0.0531$ \\
\bottomrule
\end{tabular}}
\end{minipage}
\hfill
\begin{minipage}[t]{\textwidth}
\captionsetup{type=table}
\captionof{table}{Ablation 4: Deductive Consistency vs. Hops}
\resizebox{\linewidth}{!}{
\begin{tabular}{lccccc}
\toprule
Hops & 1 & 2 & 3 & 4 & 5 \\
\midrule
Llama-3.3-70B & $0.8532 \pm 0.0134$ & $0.7876 \pm 0.0158$ & $0.7515 \pm 0.0127$ & $0.7075 \pm 0.0107$ & $0.6926 \pm 0.0076$ \\
DeepSeek-R1-Distill-Llama-70B & $0.8366 \pm 0.0238$ & $0.7726 \pm 0.0333$ & $0.7093 \pm 0.0389$ & $0.6741 \pm 0.0385$ & $0.6369 \pm 0.0307$ \\
\bottomrule
\end{tabular}}
\end{minipage}

\vspace{1em}
\end{figure}
\begin{figure}[h]
    \centering
    \begin{minipage}[t]{0.32\textwidth}
        \includegraphics[width=\linewidth]{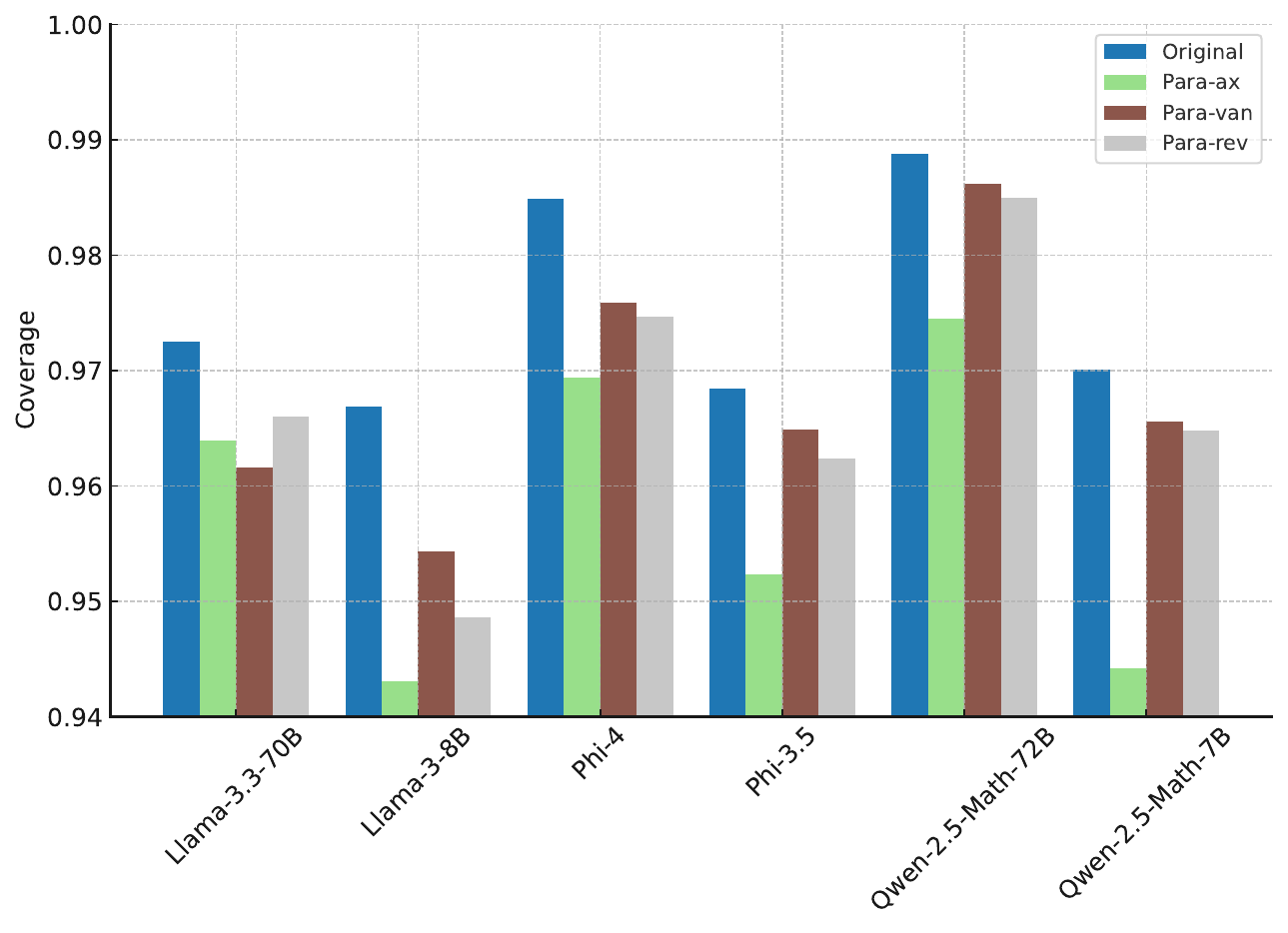}
        \captionof{figure}{Coverage Comparison Across Interventions (Adjusted Y-Limits)}
        \label{fig:paraInt}
    \end{minipage}
    \hfill
    \begin{minipage}[t]{0.32\textwidth}
        \includegraphics[width=\linewidth]{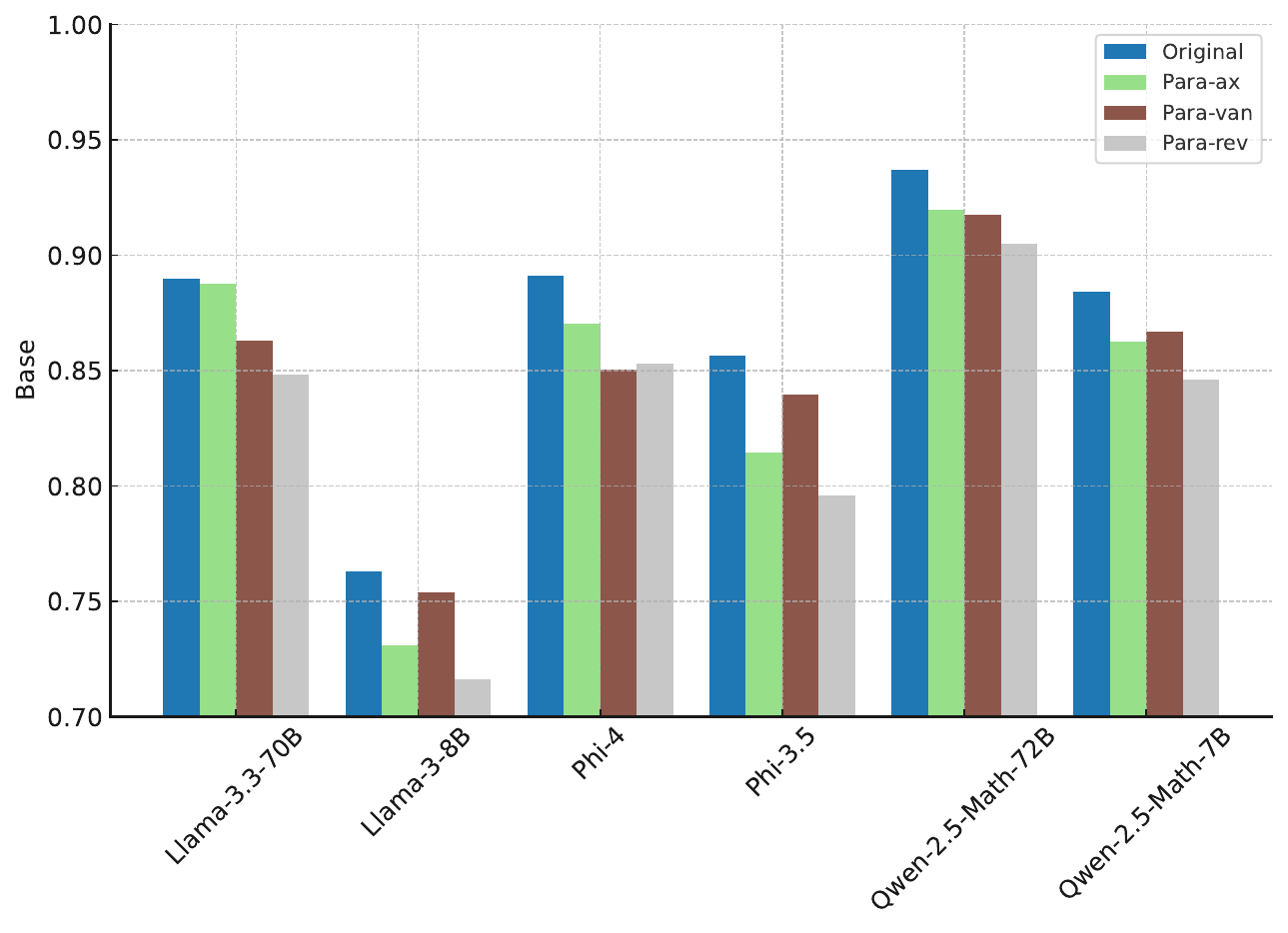}
        \captionof{figure}{Base Comparison Across Interventions}
        \label{fig:base_comparison}
    \end{minipage}
    \hfill
    \begin{minipage}[t]{0.32\textwidth}
        \includegraphics[width=\linewidth]{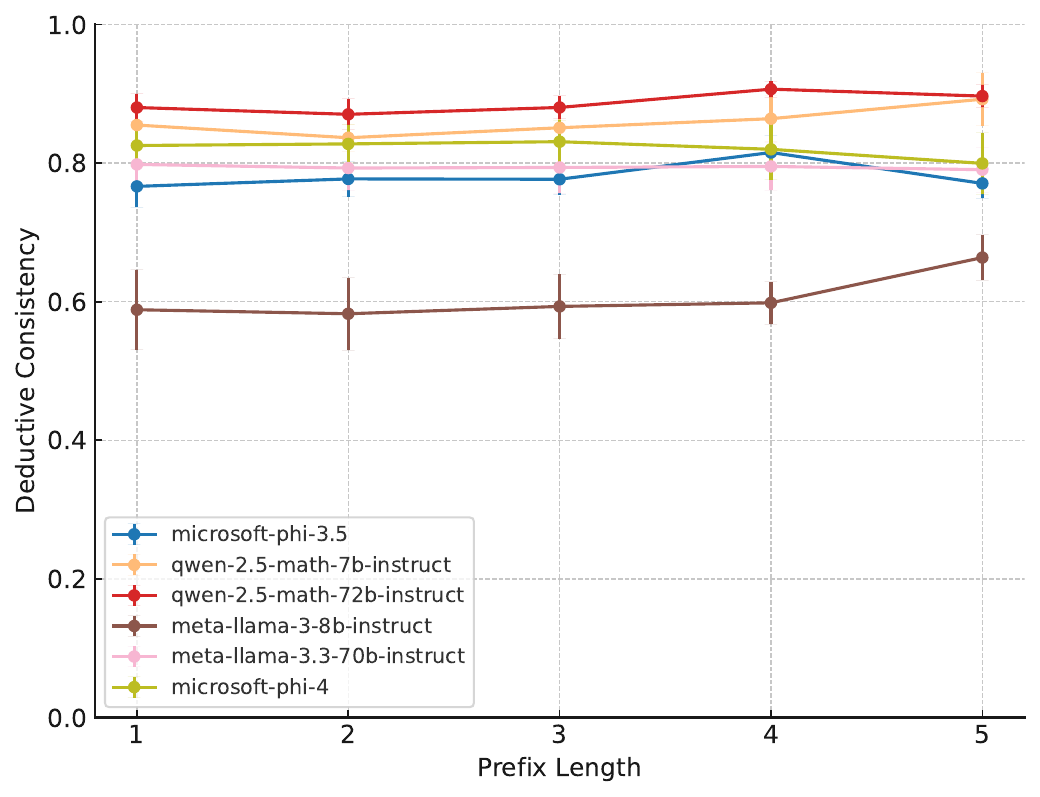}
        \captionof{figure}{Deductive Consistency vs. Prefix Length for Different Models}
        \label{fig:deductive_consistency_prefix}
    \end{minipage}
\end{figure}





\begin{figure}[!htbp]
    \centering
    \begin{subfigure}{0.24\textwidth}
        \includegraphics[width=\linewidth]{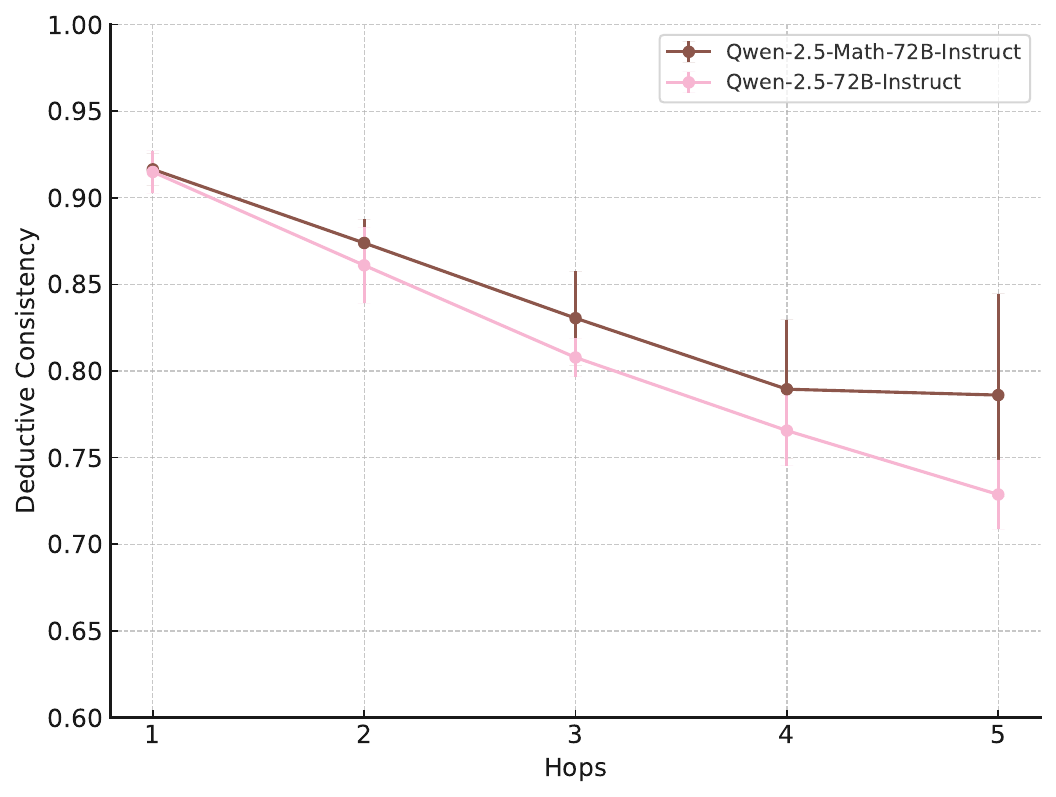}
        \caption{Ablation 1}
        \label{fig:deductive_consistency_ablation_1}
    \end{subfigure}
    \hfill
    \begin{subfigure}{0.24\textwidth}
        \includegraphics[width=\linewidth]{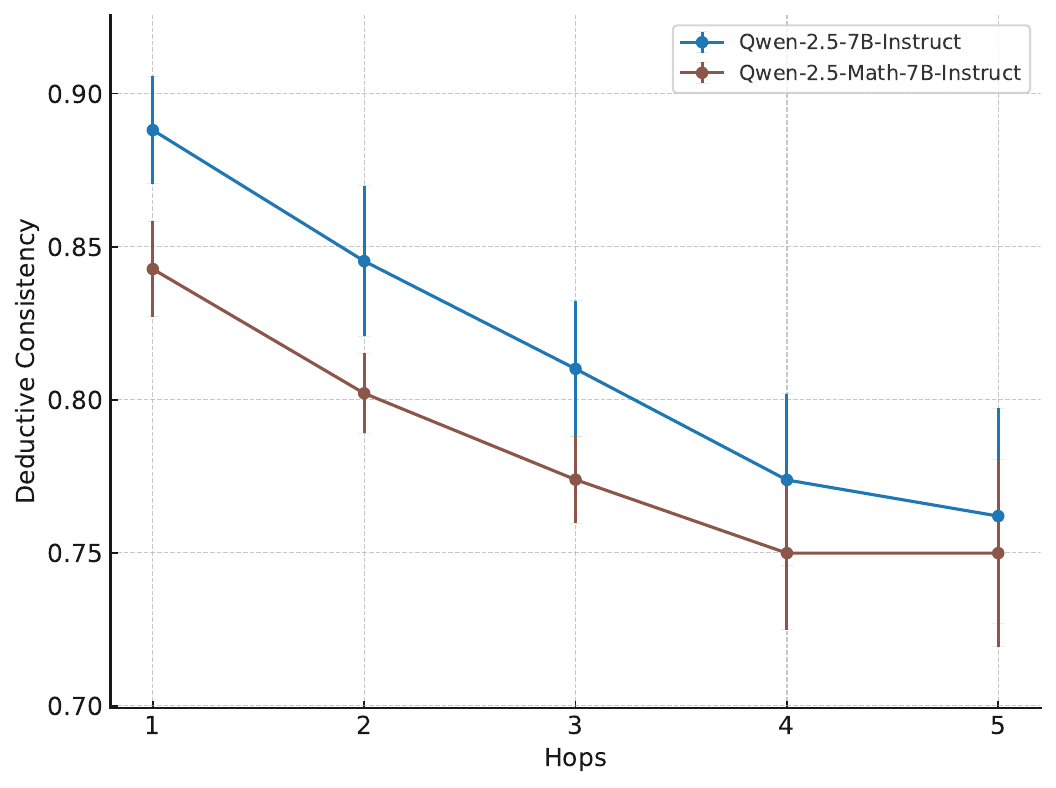}
        \caption{Ablation 2}
        \label{fig:deductive_consistency_ablation_2}
    \end{subfigure}
    \hfill
    \begin{subfigure}{0.24\textwidth}
        \includegraphics[width=\linewidth]{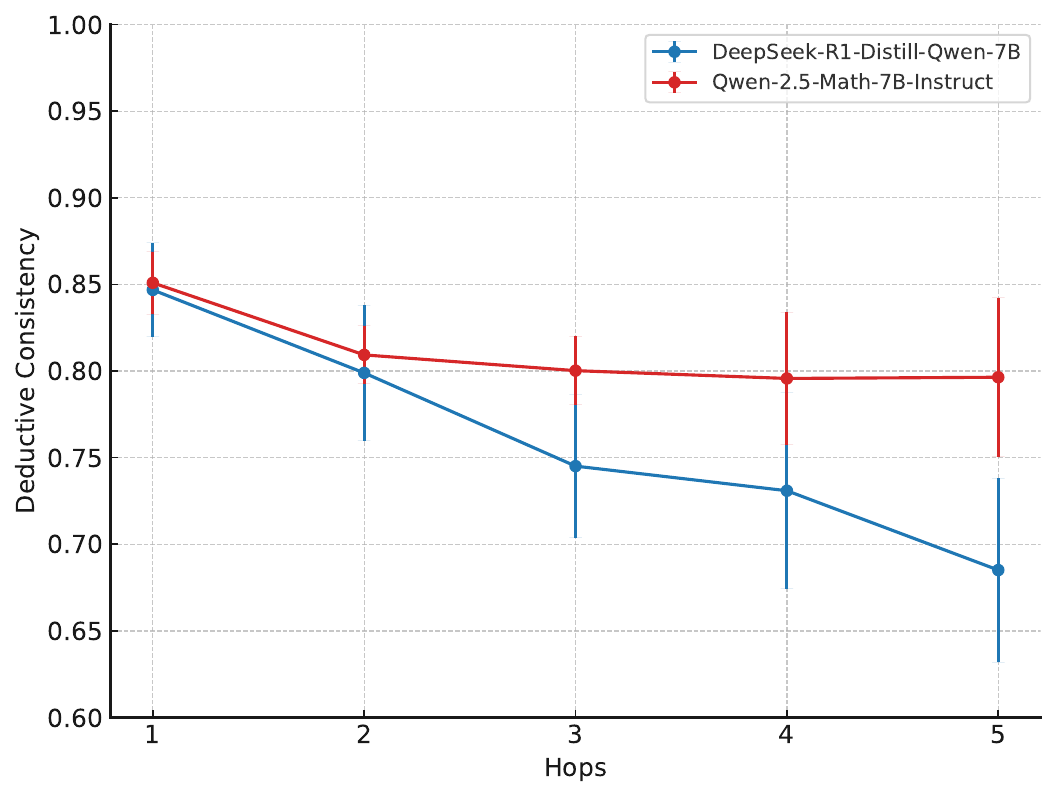}
        \caption{Ablation 3}
        \label{fig:deductive_consistency_ablation_3}
    \end{subfigure}
    \hfill
    \begin{subfigure}{0.24\textwidth}
        \includegraphics[width=\linewidth]{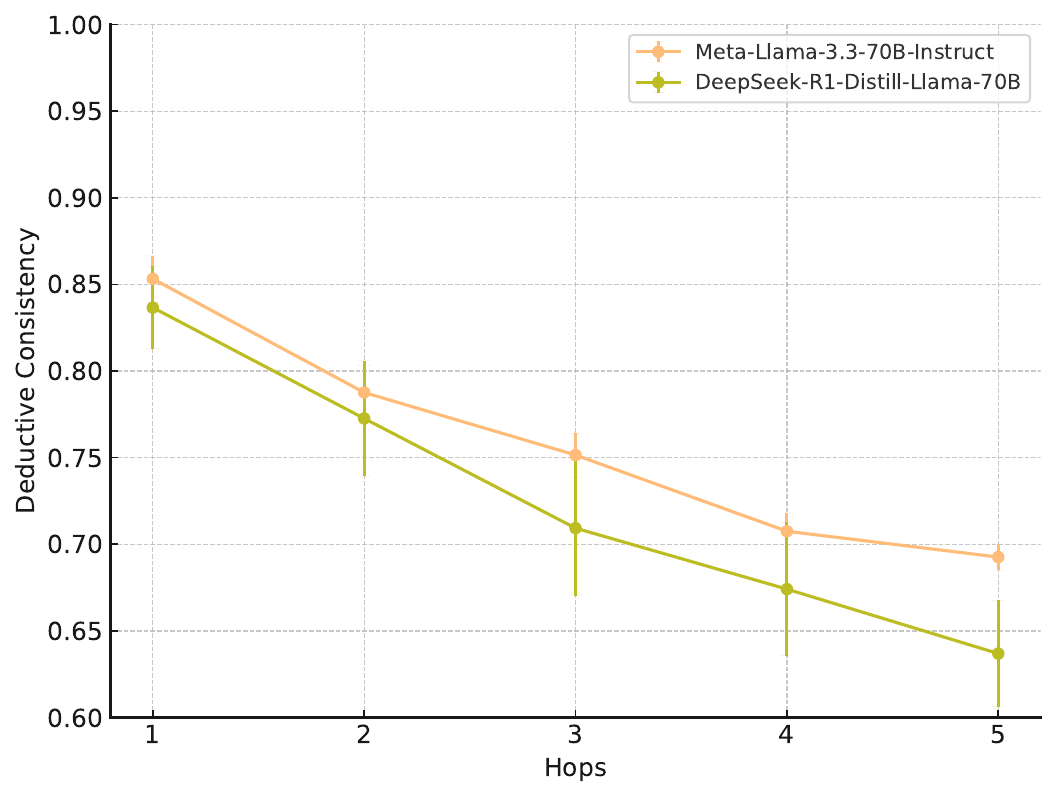}
        \caption{Ablation 4}
        \label{fig:deductive_consistency_ablation_4}
    \end{subfigure}
    \caption{Deductive Consistency vs. Hops for all Ablations}
    \label{fig:deductive_consistency_all}
\end{figure}

        


\begin{figure}[h]
    \centering
    \begin{subfigure}{0.23\textwidth}
        \includegraphics[width=\textwidth]{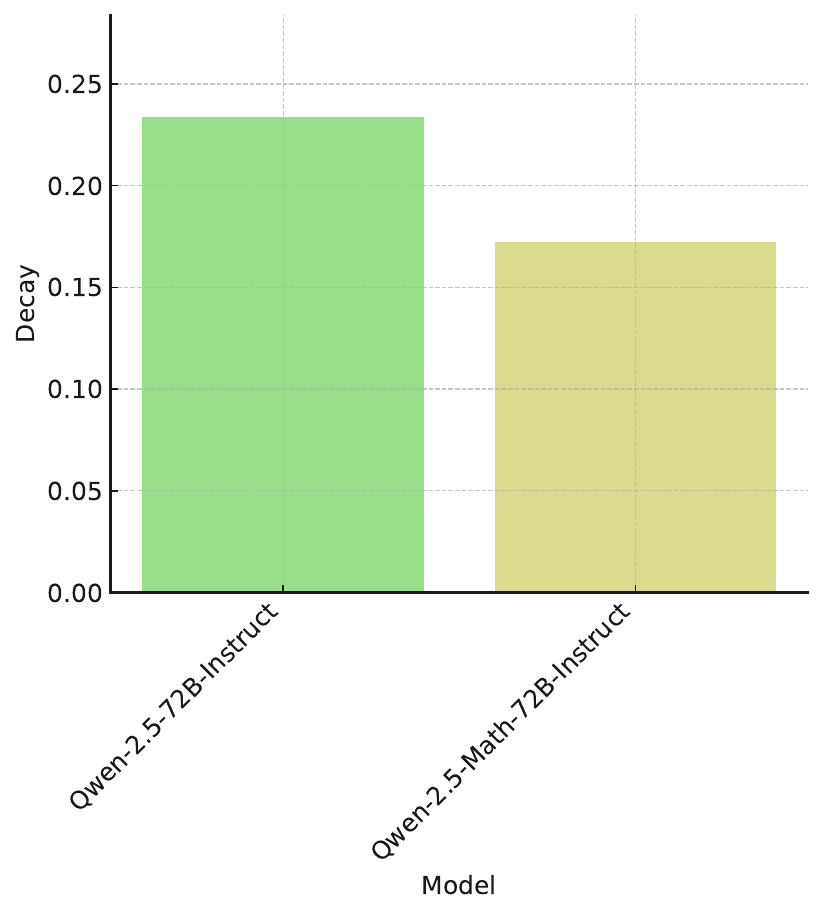}
        \caption{Ablation 1}
        \label{fig:decay_ablation_1}
    \end{subfigure}
    \hfill
    \begin{subfigure}{0.23\textwidth}
        \includegraphics[width=\textwidth]{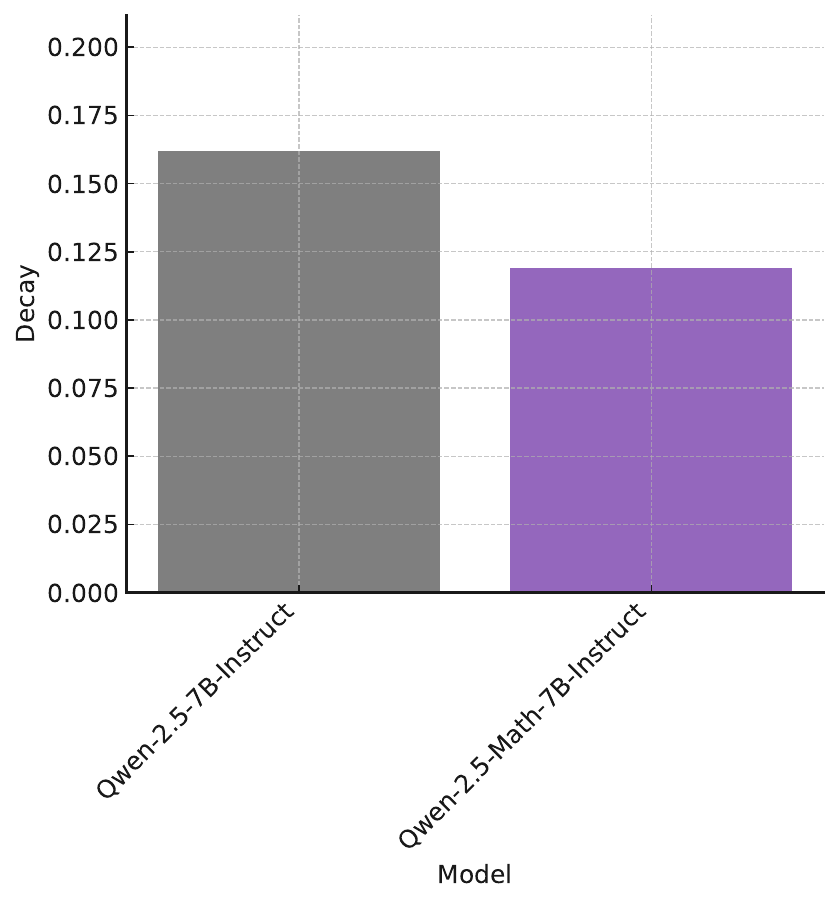}
        \caption{Ablation 2}
        \label{fig:decay_ablation_2}
    \end{subfigure}
    \hfill
    \begin{subfigure}{0.23\textwidth}
        \includegraphics[width=\textwidth]{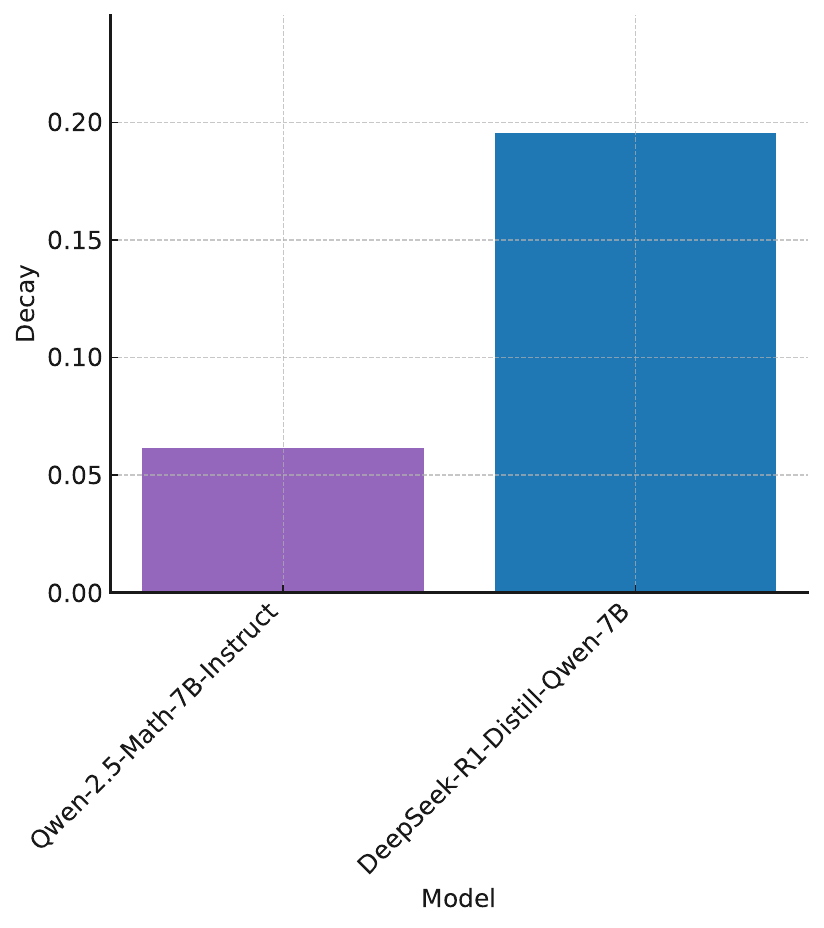}
        \caption{Ablation 3}
        \label{fig:decay_ablation_3}
    \end{subfigure}
    \hfill
    \begin{subfigure}{0.23\textwidth}
        \includegraphics[width=\textwidth]{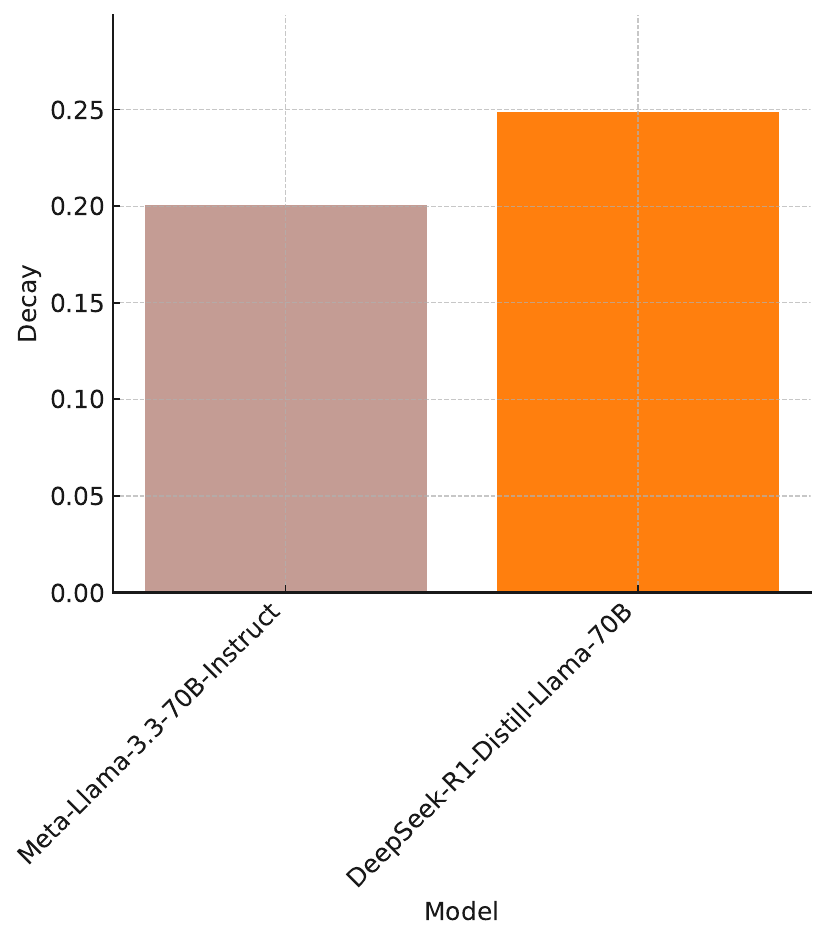}
        \caption{Ablation 4}
        \label{fig:decay_ablation_4}
    \end{subfigure}
    \caption{Comparison of Decay across Models for Ablations 1--4}
    \label{fig:decay_ablation_all}
\end{figure}

\subsection{Error Analysis on GSM8K}
\label{App:gsm8kerror}
In this section, we perform a qualitative and quantitative analysis of errors that arise in the response models.

\subsubsection{Qualitative analysis of error}
\textbf{Logical Error}
\definecolor{lightblue}{RGB}{173, 216, 230}
\definecolor{lightgray}{RGB}{240, 240, 240}
\definecolor{red}{RGB}{220, 20, 60}

\begin{tcolorbox}[colback=lightblue!15, colframe=blue!60, title=Question, rounded corners]
\textbf{Two vampires have come to the sleepy village of Willowton, population 33. Every night, each vampire turns 95 more people into vampires. How many vampires are there after two nights?}
\end{tcolorbox}

\begin{tcolorbox}[colback=lightgray!10, colframe=black!70, title=Answer (Qwen-2.5-Math-72B), rounded corners]
\textbf{Step-by-step solution:}

\textbf{Initial Condition:}  
- There are initially 2 vampires in Willowton.  

\textbf{Night 1:}  
- Each vampire turns 95 people into vampires.  
- Since there are 2 vampires, the total number of new vampires created is:  
  \[
  2 \times 95 = 190
  \]
- Total number of vampires after the first night:  
  \[
  2 + 190 = 192
  \]

\textbf{Night 2:}  
- Now there are 192 vampires.  
- Each vampire turns 95 people into vampires.  
- Since there are 192 vampires, the total number of new vampires created is:  
  \[
  192 \times 95 = 18240
  \]
- Total number of vampires after the second night:  
  \[
  192 + 18240 = 18432
  \]

\textbf{Final Answer:}  
\[
\boxed{18432}
\]
\end{tcolorbox}

\begin{tcolorbox}[colback=lightgray!15, colframe=black!70, title=Evaluation (GPT-4o), rounded corners]
\textbf{Errors:}

\begin{itemize}
    \item \textcolor{red}{\textbf{Wrong Assumption:}} The response assumes there are enough people to turn into vampires.  
    \textbf{Justification:} The village only has 33 people initially, so the vampires cannot turn more people into vampires than the village's population allows. The response fails to consider this limitation.
    
    \item \textcolor{red}{\textbf{Wrong Logic:}} The response calculates the number of new vampires incorrectly on the second night.  
    \textbf{Justification:} On the second night, the vampires cannot turn more than the remaining population into vampires. Given the initial population, they cannot create 18,240 new vampires. \textcolor{blue}{[This is logic error !!]}
\end{itemize}

\end{tcolorbox}

\textbf{Understanding Error}
\definecolor{lightblue}{RGB}{173, 216, 230}
\definecolor{lightgray}{RGB}{240, 240, 240}
\definecolor{red}{RGB}{220, 20, 60}

\begin{tcolorbox}[colback=lightblue!15, colframe=blue!60, title=Question, rounded corners]
\textbf{James writes a 36-page letter to 71 different friends 32 times a week. How many pages does he write in a year?}
\end{tcolorbox}

\begin{tcolorbox}[colback=lightgray!10, colframe=black!70, title=Answer (Phi-4), rounded corners]
\textbf{Step-by-step solution:}

1. \textbf{Calculate the total number of pages written per week:}  
   - James writes a 36-page letter to each of his 71 friends.  
   - Total pages per week:
   \[
   \text{Total pages per week} = 36 \times 71 = 2,556 \text{ pages}
   \]

2. \textbf{Calculate the total number of pages written in a year:}  
   - There are 52 weeks in a year.
   \[
   \text{Total pages per year} = 2,556 \times 52 = 132,912 \text{ pages}
   \]

\textbf{Final Answer:}  
\[
\boxed{132,912} \text{ pages}
\]

\end{tcolorbox}

\begin{tcolorbox}[colback=lightgray!15, colframe=black!70, title=Evaluation (GPT-4o), rounded corners]
\textbf{Errors:}

\begin{itemize}
    \item \textcolor{red}{\textbf{Wrong Assumption:}} The response assumes James writes a 36-page letter to each friend only once per week.  
    \textbf{Justification:} The problem states he writes 32 times per week, so the calculation of total pages per week is incorrect.

    \item \textcolor{red}{\textbf{Calculation Error:}} The response calculates the total number of pages per year using an incorrect assumption.  
    \textbf{Justification:} Since the number of times he writes per week is not factored in, the yearly calculation does not reflect the correct total.
\end{itemize}

\end{tcolorbox}

\textbf{Calculation Error}

\definecolor{lightblue}{RGB}{173, 216, 230}
\definecolor{lightgray}{RGB}{240, 240, 240}
\definecolor{red}{RGB}{220, 20, 60}

\begin{tcolorbox}[colback=lightblue!15, colframe=blue!60, title=Question, rounded corners]
\textbf{Max likes to collect model trains. He asks for 53 for every birthday of his, and asks for 18 each Christmas. Max always gets the gifts he asks for, and asks for these same gifts every year for 16 years. At the end of the 16 years, his parents give him double the number of trains he already has. How many trains does Max have now?}
\end{tcolorbox}

\begin{tcolorbox}[colback=lightgray!10, colframe=black!70, title=Answer (Llama-3-8B), rounded corners]
\textbf{Step-by-step solution:}

1. Max gets 53 trains for each birthday, and he has 16 birthdays:  
   \[
   53 \times 16 = 848 \text{ trains}
   \]

2. Max gets 18 trains for each Christmas, and he has 16 Christmases:  
   \[
   18 \times 16 = 288 \text{ trains}
   \]

3. In total, Max receives:  
   \[
   848 + 288 = 1,136 \text{ trains}
   \]

4. At the end of the 16 years, his parents give him double the number of trains he already has:  
   \[
   1,136 \times 2 = 2,268 \text{ trains}
   \]

5. In total, Max now has:  
   \[
   1,136 + 2,268 = 3,404 \text{ trains}
   \]

\textbf{Final Answer:}  
\[
\boxed{3,404} \text{ trains}
\]

\end{tcolorbox}

\begin{tcolorbox}[colback=lightgray!15, colframe=black!70, title=Evaluation (GPT-4o), rounded corners]
\textbf{Errors:}

\begin{itemize}
    \item \textcolor{red}{\textbf{Calculation Error:}} The response states that at the end of the 16 years, Max's parents give him double the number of trains he already has, calculated as \( 1,136 \times 2 = 2,268 \).  
    \textbf{Justification:} The correct calculation should be:
    \[
    1,136 \times 2 = 2,272
    \]

    \item \textcolor{red}{\textbf{Propagated Error Only:}} The response states that the total number of trains Max has is \( 1,136 + 2,268 = 3,404 \).  
    \textbf{Justification:} Since the previous multiplication was incorrect, the final sum was also incorrect. The correct final total should be:
    \[
    1,136 + 2,272 = 3,408
    \]
\end{itemize}

\end{tcolorbox}

\subsubsection{Quantitative analysis of error}
\begin{table}[h!]
    \centering
    \renewcommand{\arraystretch}{1.2}
    \begin{tabular}{lccccc}
        \toprule
        \textbf{Models} & \textbf{Group 1} & \textbf{Group 2} & \textbf{Group 3} & \textbf{Group 4} & \textbf{Group 5} \\
        \midrule
        \textbf{Llama-3-8B} & 44 & 50 & 252 & 412 & 180 \\
        \textbf{Llama-3.3-70B} & 92 & 90 & 143 & 209 & 90 \\
        \textbf{Qwen-2.5-Math-72B} & 916 & 338 & 139 & 170 & 60 \\
        \textbf{Qwen-2.5-Math-7B} & 560 & 450 & 240 & 240 & 110 \\
        \textbf{Phi-4} & 553 & 117 & 131 & 199 & 130 \\
        \textbf{Phi-3.5} & 77 & 139 & 125 & 255 & 170 \\
        \bottomrule
    \end{tabular}
    \caption{Number of instances for different models across dataset groups.}
    \label{tab:instances_per_group}
\end{table}

\begin{figure}[!htbp]
    \centering
    \begin{subfigure}[b]{0.45\textwidth}
        \includegraphics[width=\textwidth]{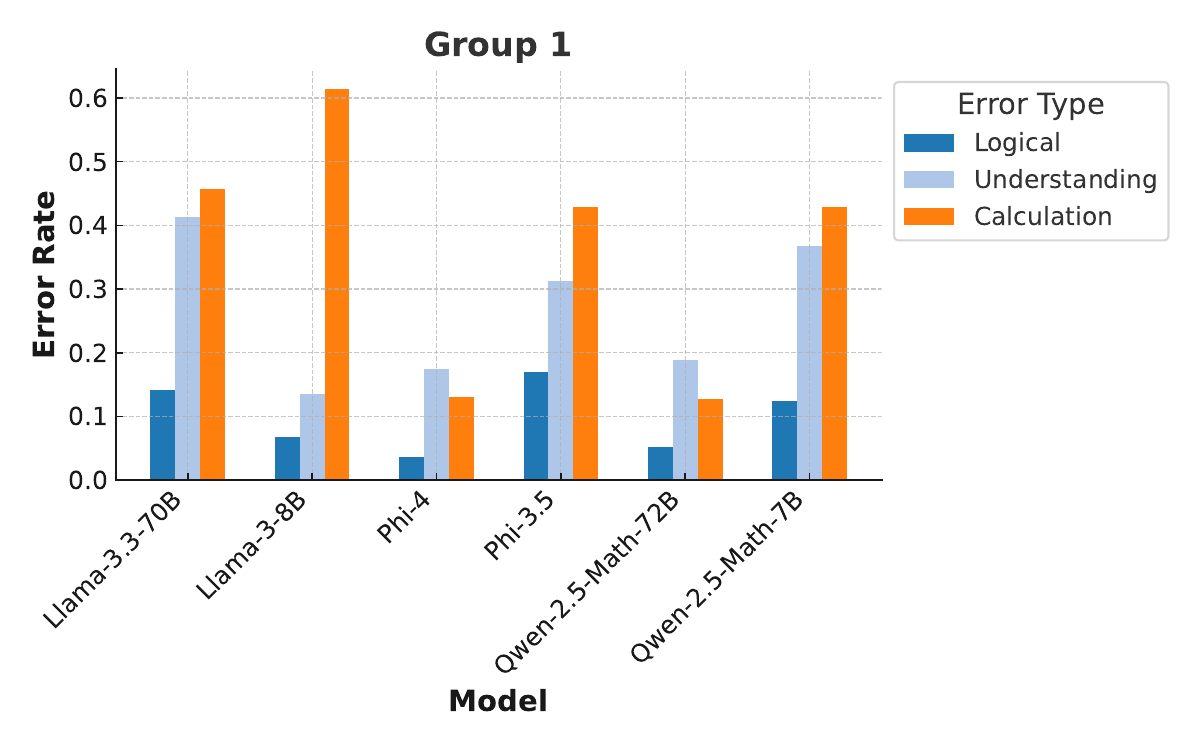}
        \caption{Group 1}
        \label{fig:error_group_1}
    \end{subfigure}
    \hfill
    \begin{subfigure}[b]{0.45\textwidth}
        \includegraphics[width=\textwidth]{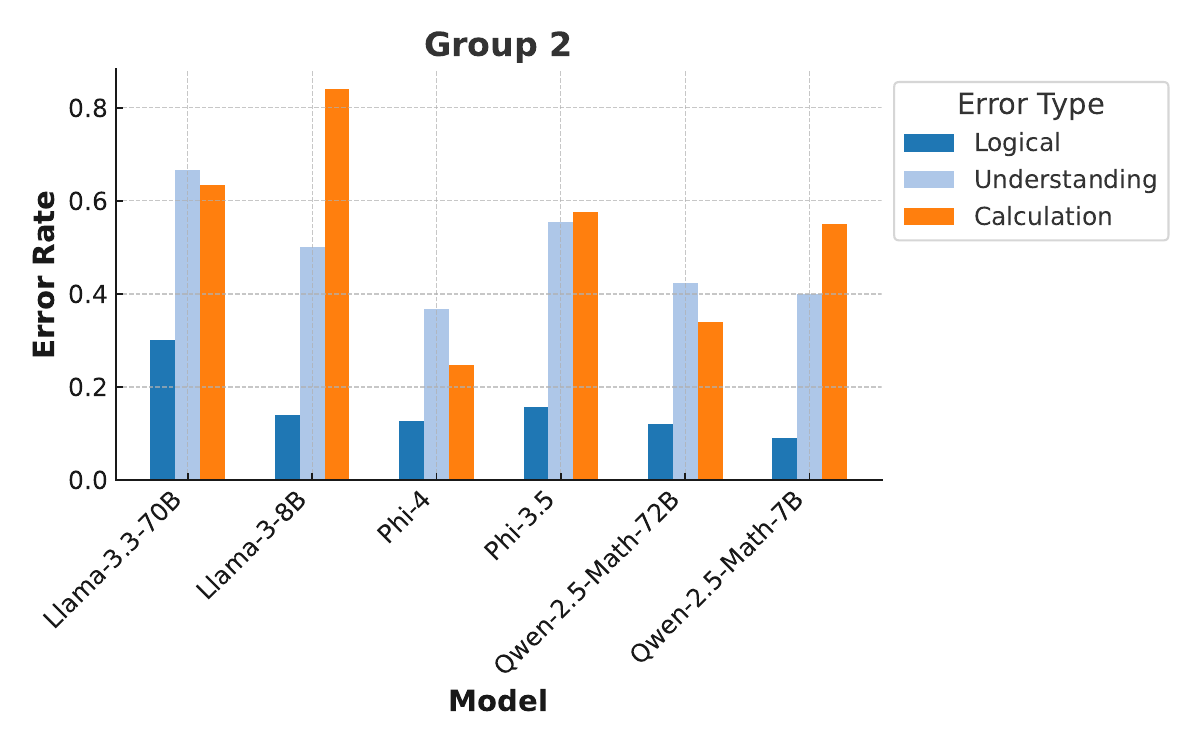}
        \caption{Group 2}
        \label{fig:error_group_2}
    \end{subfigure}

    \vspace{1em}

    \begin{subfigure}[b]{0.45\textwidth}
        \includegraphics[width=\textwidth]{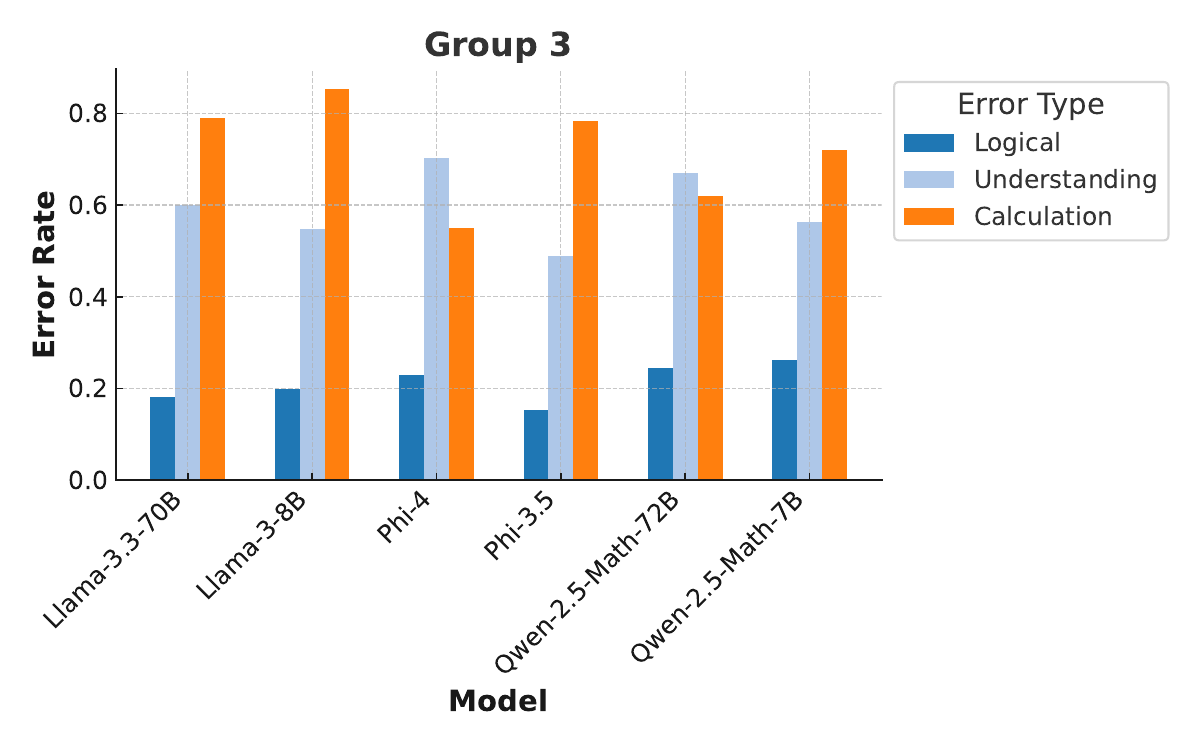}
        \caption{Group 3}
        \label{fig:error_group_3}
    \end{subfigure}
    \hfill
    \begin{subfigure}[b]{0.45\textwidth}
        \includegraphics[width=\textwidth]{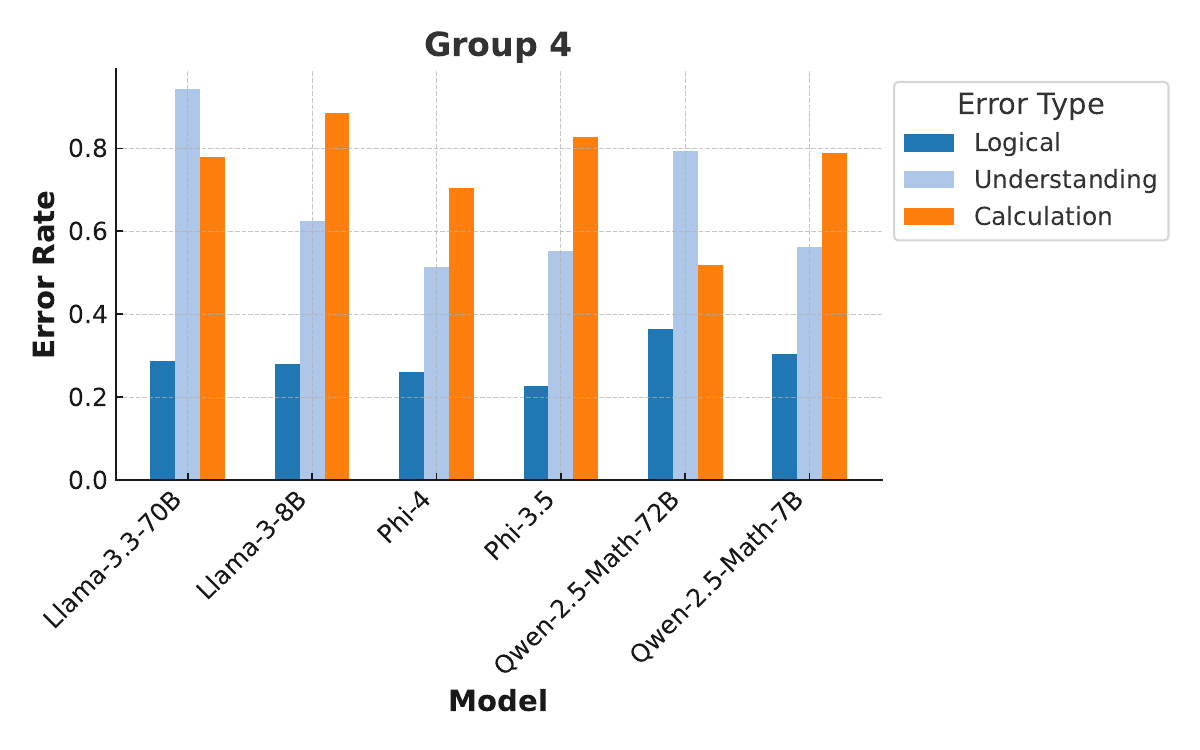}
        \caption{Group 4}
        \label{fig:error_group_4}
    \end{subfigure}

    \vspace{1em}

    \begin{subfigure}[b]{0.45\textwidth}
        \includegraphics[width=\textwidth]{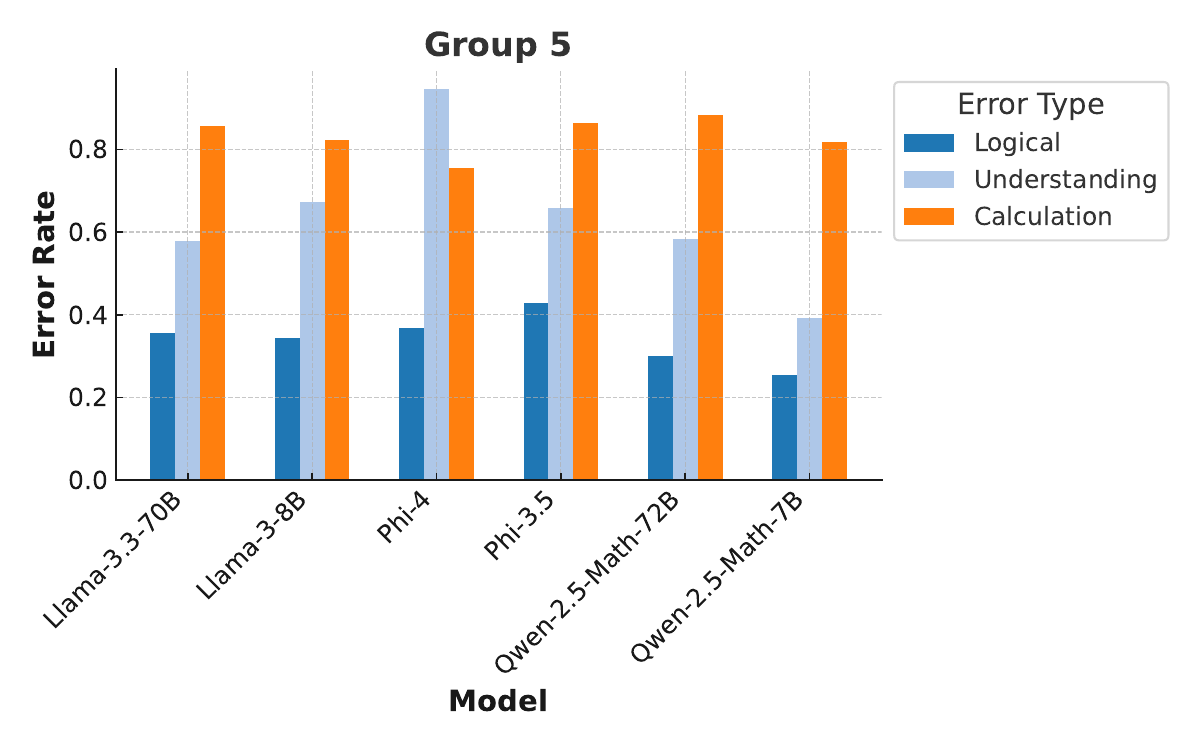}
        \caption{Group 5}
        \label{fig:error_group_5}
    \end{subfigure}

    \caption{Error rate plots for all groups}
    \label{fig:error_all_groups}
\end{figure}





\clearpage
\subsection{SynDeduct}
\label{sub:synRes}
\begin{table}[h]
    \centering
    \caption{DC ± Standard Error for Hops 1-6}
    \resizebox{\textwidth}{!}{%
    \begin{tabular}{lcccccc}
        \toprule
        \textbf{Model} & \textbf{Hop 1} & \textbf{Hop 2} & \textbf{Hop 3} & \textbf{Hop 4} & \textbf{Hop 5} & \textbf{Hop 6} \\
        \midrule
        Qwen-2.5-Math-7B & $0.2083 \pm 0.0336$ & $0.1205 \pm 0.0354$ & $0.0867 \pm 0.0228$ & $0.0898 \pm 0.0245$ & $0.0792 \pm 0.0238$ & $0.0607 \pm 0.0189$ \\
        Qwen-2.5-7B & $0.5458 \pm 0.0376$ & $0.3705 \pm 0.0474$ & $0.3250 \pm 0.0430$ & $0.2630 \pm 0.0357$ & $0.2562 \pm 0.0279$ & $0.2440 \pm 0.0334$ \\
        Qwen-2.5-Math-72B & $0.5674 \pm 0.0285$ & $0.4894 \pm 0.0434$ & $0.4433 \pm 0.0456$ & $0.3852 \pm 0.0553$ & $0.3635 \pm 0.0513$ & $0.3381 \pm 0.0549$ \\
        Qwen-2.5-72B & $0.6868 \pm 0.0287$ & $0.5848 \pm 0.0389$ & $0.4825 \pm 0.0368$ & $0.4046 \pm 0.0358$ & $0.3354 \pm 0.0294$ & $0.2643 \pm 0.0215$ \\
        Llama-3-8B & $0.2993 \pm 0.0458$ & $0.2023 \pm 0.0411$ & $0.1825 \pm 0.0382$ & $0.1602 \pm 0.0368$ & $0.1469 \pm 0.0275$ & $0.1357 \pm 0.0257$ \\
        DeepSeek-R1-Llama-70B & $0.7389 \pm 0.0202$ & $0.6879 \pm 0.0196$ & $0.6742 \pm 0.0175$ & $0.6509 \pm 0.0169$ & $0.6542 \pm 0.0167$ & $0.6488 \pm 0.0206$ \\
        Llama-3.3-70B & $0.8465 \pm 0.0124$ & $0.8129 \pm 0.0158$ & $0.7675 \pm 0.0147$ & $0.7250 \pm 0.0140$ & $0.7125 \pm 0.0177$ & $0.6833 \pm 0.0096$ \\
        DeepSeek-R1-Qwen-7B & $0.5424 \pm 0.0369$ & $0.3871 \pm 0.0442$ & $0.3308 \pm 0.0416$ & $0.2870 \pm 0.0432$ & $0.2802 \pm 0.0441$ & $0.2262 \pm 0.0348$ \\
        \bottomrule
    \end{tabular}%
    }
    \label{tab:dc_vs_hops_1_6}
\end{table}

\begin{table}[h]
    \centering
    \caption{DC ± Standard Error for Hops 7-12}
    \resizebox{\textwidth}{!}{%
    \begin{tabular}{lcccccc}
        \toprule
        \textbf{Model} & \textbf{Hop 7} & \textbf{Hop 8} & \textbf{Hop 9} & \textbf{Hop 10} & \textbf{Hop 11} & \textbf{Hop 12} \\
        \midrule
        Qwen-2.5-Math-7B & $0.0403 \pm 0.0105$ & $0.0383 \pm 0.0124$ & $0.0250 \pm 0.0088$ & $0.0250 \pm 0.0097$ & $0.0167 \pm 0.0090$ & $0.0000 \pm 0.0000$ \\
        Qwen-2.5-7B & $0.1639 \pm 0.0196$ & $0.1633 \pm 0.0217$ & $0.1479 \pm 0.0256$ & $0.0917 \pm 0.0273$ & $0.0792 \pm 0.0232$ & $0.0500 \pm 0.0186$ \\
        Qwen-2.5-Math-72B & $0.3222 \pm 0.0582$ & $0.3367 \pm 0.0578$ & $0.3229 \pm 0.0678$ & $0.2861 \pm 0.0648$ & $0.2667 \pm 0.0691$ & $0.2583 \pm 0.0702$ \\
        Qwen-2.5-72B & $0.2167 \pm 0.0176$ & $0.1750 \pm 0.0164$ & $0.1562 \pm 0.0209$ & $0.0889 \pm 0.0149$ & $0.0667 \pm 0.0136$ & $0.0417 \pm 0.0185$ \\
        Llama-3-8B & $0.0903 \pm 0.0215$ & $0.0883 \pm 0.0201$ & $0.1104 \pm 0.0229$ & $0.0861 \pm 0.0169$ & $0.0417 \pm 0.0154$ & $0.0083 \pm 0.0080$ \\
        DeepSeek-R1-Llama-70B & $0.6431 \pm 0.0159$ & $0.6083 \pm 0.0248$ & $0.5958 \pm 0.0273$ & $0.5556 \pm 0.0186$ & $0.4667 \pm 0.0325$ & $0.3333 \pm 0.0430$ \\
        Llama-3.3-70B & $0.6347 \pm 0.0156$ & $0.6017 \pm 0.0169$ & $0.5854 \pm 0.0178$ & $0.5750 \pm 0.0391$ & $0.5417 \pm 0.0316$ & $0.6083 \pm 0.0343$ \\
        DeepSeek-R1-Qwen-7B & $0.2097 \pm 0.0336$ & $0.1633 \pm 0.0360$ & $0.1604 \pm 0.0353$ & $0.1250 \pm 0.0243$ & $0.0667 \pm 0.0198$ & $0.0667 \pm 0.0136$ \\
        \bottomrule
    \end{tabular}
    }
    \label{tab:dc_vs_hops_7_12}
\end{table}

\begin{table}[h]
    \centering
    \caption{DC ± Standard Error for Prefix 1-6}
    \resizebox{\textwidth}{!}{%
    \begin{tabular}{lcccccc}
        \toprule
        \textbf{Model} & \textbf{Prefix 1} & \textbf{Prefix 2} & \textbf{Prefix 3} & \textbf{Prefix} & \textbf{Prefix 5} & \textbf{Prefix 6} \\
        \midrule
        Qwen-2.5-Math-7B & $0.2113 \pm 0.0452$ & $0.1118 \pm 0.0257$ & $0.0967 \pm 0.0218$ & $0.0878 \pm 0.0231$ & $0.0663 \pm 0.0214$ & $0.0734 \pm 0.0132$ \\
        Qwen-2.5-7B & $0.3719 \pm 0.0678$ & $0.3715 \pm 0.0483$ & $0.2947 \pm 0.0559$ & $0.2951 \pm 0.0452$ & $0.2289 \pm 0.0479$ & $0.2401 \pm 0.0421$ \\
        Qwen-2.5-Math-72B & $0.7968 \pm 0.0140$ & $0.6433 \pm 0.0187$ & $0.4900 \pm 0.0273$ & $0.3836 \pm 0.0338$ & $0.3812 \pm 0.0313$ & $0.3475 \pm 0.0289$ \\
        Qwen-2.5-72B & $0.3674 \pm 0.0737$ & $0.3526 \pm 0.0773$ & $0.3674 \pm 0.0714$ & $0.3522 \pm 0.0777$ & $0.3039 \pm 0.0670$ & $0.3396 \pm 0.0545$ \\
        Llama-3-8B & $0.3011 \pm 0.0451$ & $0.2580 \pm 0.0341$ & $0.2371 \pm 0.0430$ & $0.1494 \pm 0.0285$ & $0.1169 \pm 0.0277$ & $0.0981 \pm 0.0191$ \\
        DeepSeek-R1-Llama-70B & $0.5970 \pm 0.0546$ & $0.6546 \pm 0.0570$ & $0.6828 \pm 0.0401$ & $0.6461 \pm 0.0569$ & $0.6287 \pm 0.0563$ & $0.6304 \pm 0.0474$ \\
        Llama-3.3-70B & $0.6452 \pm 0.0204$ & $0.7562 \pm 0.0186$ & $0.6720 \pm 0.0298$ & $0.6691 \pm 0.0292$ & $0.6730 \pm 0.0181$ & $0.6094 \pm 0.0301$ \\
        DeepSeek-R1-Qwen-7B & $0.4534 \pm 0.0581$ & $0.3817 \pm 0.0550$ & $0.3252 \pm 0.0552$ & $0.2780 \pm 0.0440$ & $0.3136 \pm 0.0376$ & $0.1923 \pm 0.0406$ \\
        \bottomrule
    \end{tabular}%
    }
    \label{tab:prefix_dc_vs_hops_1_6}
\end{table}

\begin{table}[h]
    \centering
    \caption{DC ± Standard Error for Prefix 7-12}
    \resizebox{\textwidth}{!}{%
    \begin{tabular}{lcccccc}
        \toprule
        \textbf{Model} & \textbf{Prefix 7} & \textbf{Prefix 8} & \textbf{Prefix} & \textbf{Prefix 10} & \textbf{Prefix 11} & \textbf{Prefix 12} \\
        \midrule
        Qwen-2.5-Math-7B & $0.0383 \pm 0.0157$ & $0.0407 \pm 0.0106$ & $0.0191 \pm 0.0080$ & $0.0151 \pm 0.0061$ & $0.0176 \pm 0.0100$ & $0.0124 \pm 0.0061$ \\
        Qwen-2.5-7B & $0.1651 \pm 0.0386$ & $0.1544 \pm 0.0383$ & $0.1918 \pm 0.0329$ & $0.1376 \pm 0.0362$ & $0.1277 \pm 0.0238$ & $0.1217 \pm 0.0227$ \\
        Qwen-2.5-Math-72B & $0.2979 \pm 0.0410$ & $0.3003 \pm 0.0274$ & $0.2063 \pm 0.0336$ & $0.1619 \pm 0.0301$ & $0.1676 \pm 0.0401$ & $0.2034 \pm 0.0359$ \\
        Qwen-2.5-72B & $0.2892 \pm 0.0583$ & $0.2505 \pm 0.0558$ & $0.2376 \pm 0.0490$ & $0.2286 \pm 0.0419$ & $0.2007 \pm 0.0436$ & $0.2138 \pm 0.0478$ \\
        Llama-3-8B & $0.0775 \pm 0.0240$ & $0.0903 \pm 0.0147$ & $0.0821 \pm 0.0163$ & $0.0362 \pm 0.0085$ & $0.0599 \pm 0.0091$ & $0.0452 \pm 0.0108$ \\
        DeepSeek-R1-Llama-70B & $0.6311 \pm 0.0498$ & $0.6519 \pm 0.0377$ & $0.6133 \pm 0.0316$ & $0.6845 \pm 0.0149$ & $0.6875 \pm 0.0352$ & $0.6426 \pm 0.0473$ \\
        Llama-3.3-70B & $0.6164 \pm 0.0176$ & $0.5876 \pm 0.0226$ & $0.6538 \pm 0.0195$ & $0.5721 \pm 0.0317$ & $0.5625 \pm 0.0206$ & $0.5844 \pm 0.0122$ \\
        DeepSeek-R1-Qwen-7B & $0.1470 \pm 0.0350$ & $0.1923 \pm 0.0339$ & $0.1927 \pm 0.0377$ & $0.1070 \pm 0.0292$ & $0.1302 \pm 0.0344$ & $0.1321 \pm 0.0300$ \\
        \bottomrule
    \end{tabular}%
    }
    \label{tab:prefix_dc_vs_hops_7_12}
\end{table}

\begin{figure}[h]
    \centering
    \includegraphics[width=0.6\textwidth]{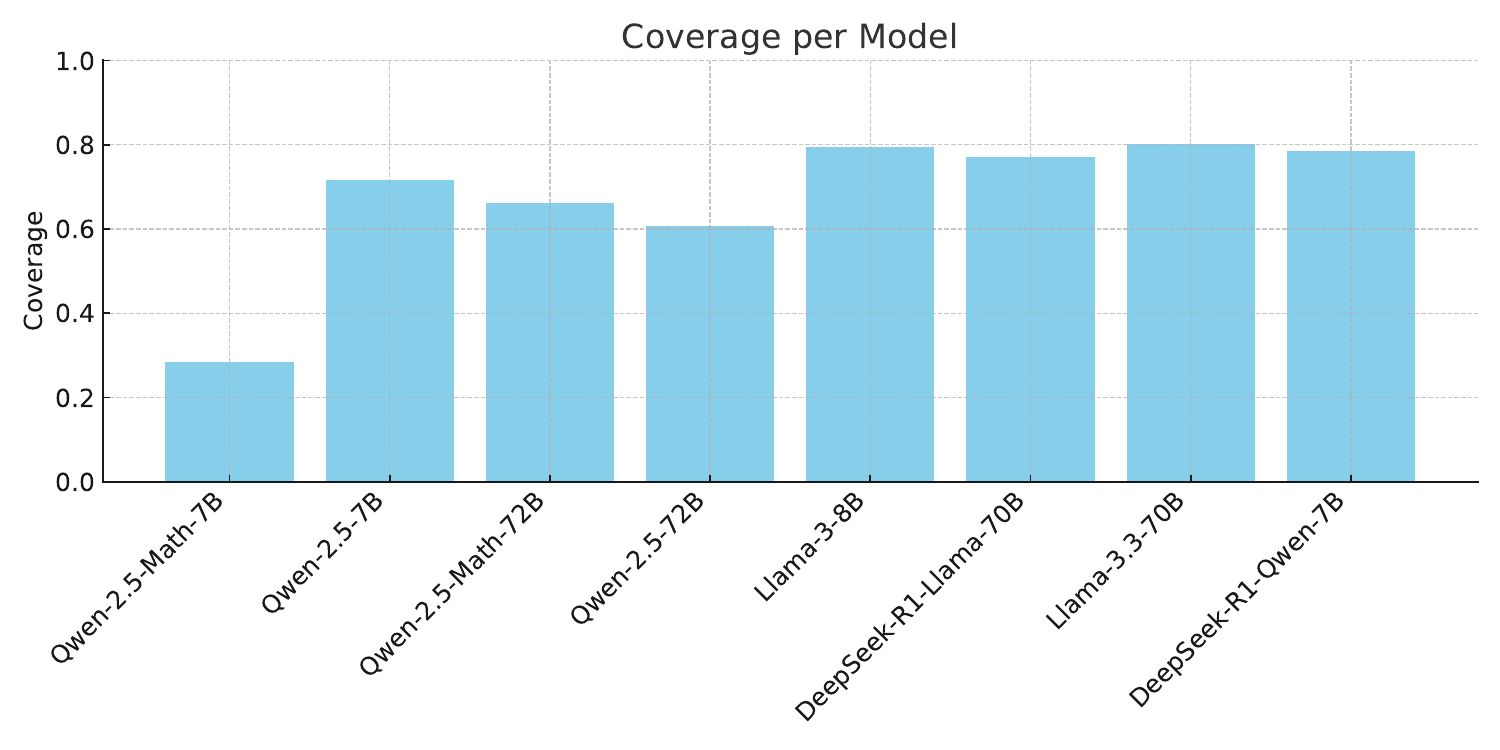}
    \caption{Coverage Metric in SynDeDeduct for Different Models}
    \label{fig:coverage_syndeduct}
\end{figure}

\begin{figure}[h]
    \centering
    \subfloat[DC vs. Hops for Group 1]{
        \includegraphics[width=0.40\textwidth]{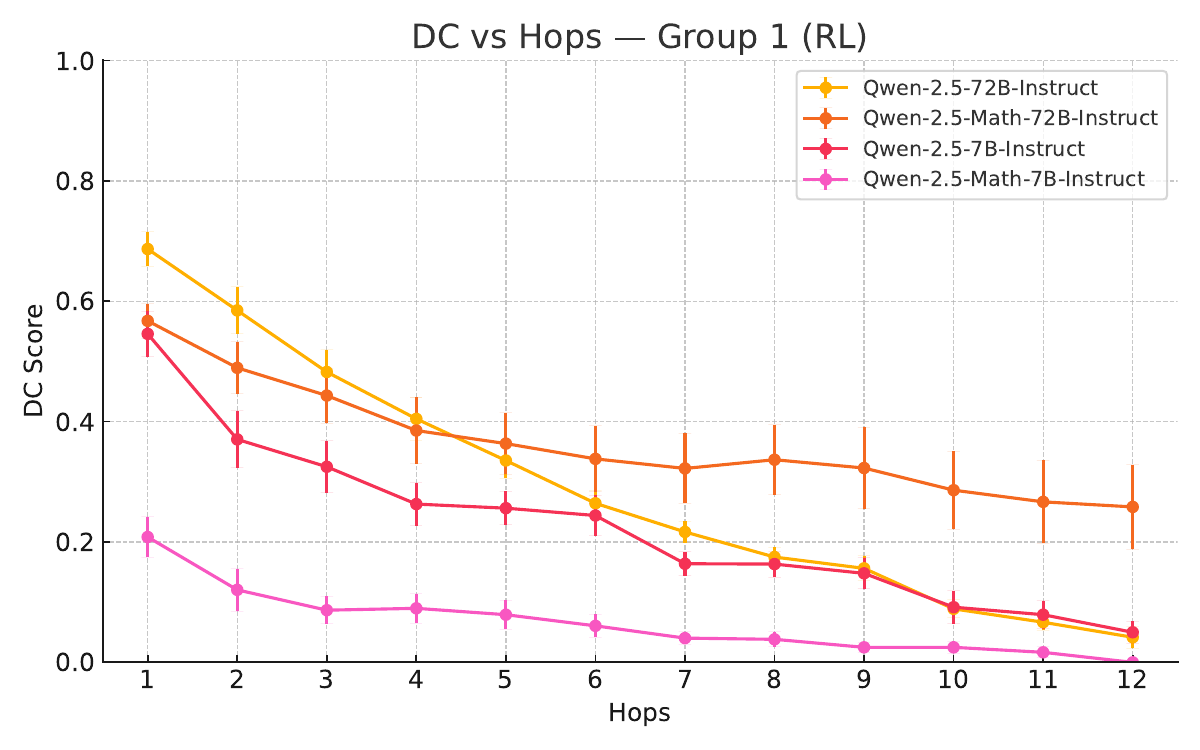}
        \label{fig:accuracy_hops_group_1}
    }
    \hfill
    \subfloat[DC vs. Hops for Group 2]{
        \includegraphics[width=0.40\textwidth]{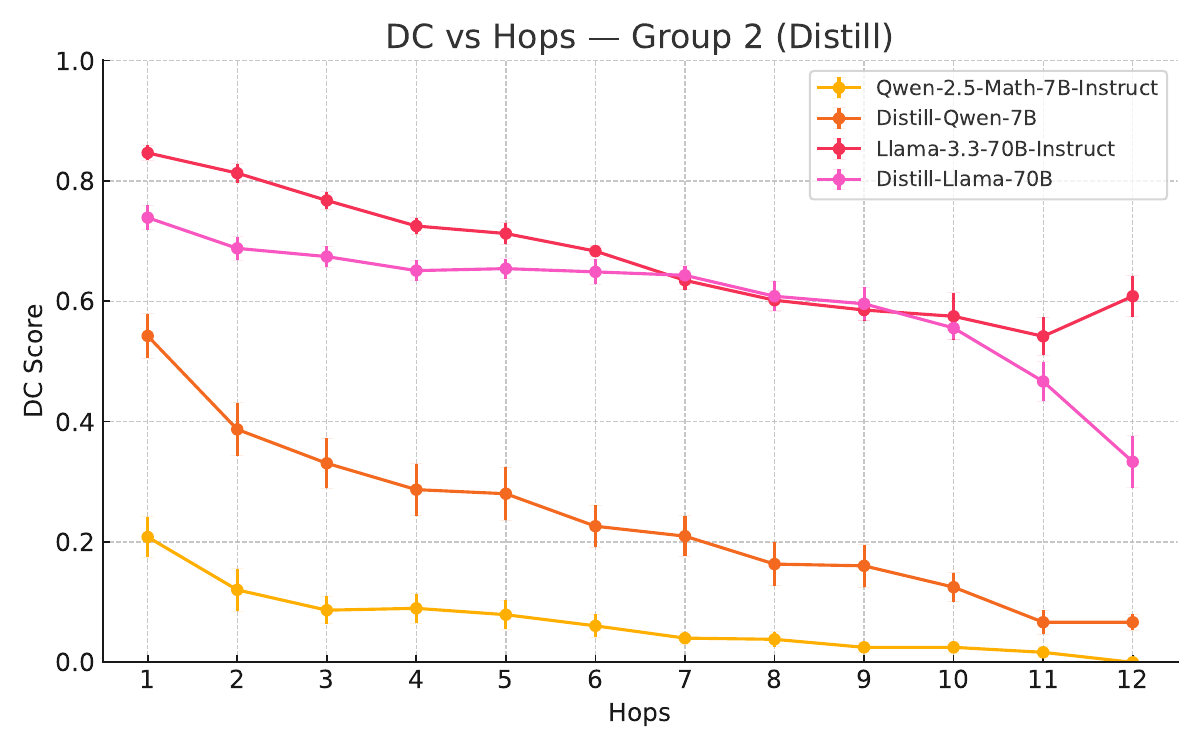}
        \label{fig:accuracy_hops_group_2}
    }
    \caption{Comparison of DC vs. Hops for Different Model Groups}
\end{figure}

\clearpage
\subsection{LM Prompts for GSM8K}
\label{App:PromptsGSM8K}

\begin{tcolorbox}[colback=blue!5!white, colframe=blue!75!black, title=LLM Prompt]
\textbf{System prompt:} "You are an expert in math. You must answer this question. You give the step-by-step solution followed by four \# symbols followed by the answer."

\textbf{User:} "Question: \textless question\textgreater \textbackslash nAnswer:"

\vspace{0.5cm}

\textit{Example Question:} \\
\textbf{User:} "Question: Yasna has two books. One book is 60 pages long, and the other book is 12 pages long. If Yasna wants to finish both of the books in 6 weeks, how many pages will Yasna need to read every day, if she reads an equal number of pages each day? \textbackslash nAnswer:"

\vspace{0.5cm}

\textit{Example Assistant Partial Answer:}

\textbf{Assistant:}
\begin{tcolorbox}[colback=red!10!white, colframe=red!75!black]
Yasna has \( 60 + 12 = 72 \) pages to read.
\end{tcolorbox}

\begin{tcolorbox}[colback=green!10!white, colframe=green!75!black]
She has \( 6 \times 7 = 42 \) days to read them.
\end{tcolorbox}

\end{tcolorbox}

\begin{tcolorbox}[colback=blue!5!white, colframe=blue!75!black, title=Template Builder Agent Prompt, breakable]

\textbf{System Prompt:} \\
You are a templatizing agent. Your task is to process questions and answers, templatize them by replacing specific numerical values with placeholders, and create a structured JSON output. The JSON output must contain the following keys:

1. **templatized\_question**: A version of the question where specific numerical values, object names, or other unique entities are replaced by placeholders.\\
2. **templatized\_answer**: A step-by-step reasoning answer where specific numerical values or entities are replaced by placeholders. Each step should remain logically consistent with the original answer.\\
3. **factual\_assignment**: A dictionary mapping placeholders to their original factual values, ensuring the templatized versions can reconstruct the original question and answer. MUST only contain NUMERICAL values.

Make sure the templatized answer and question ALIGN PERFECTLY with the original answer structure.

---

\#\#\# **ICL Examples**

\#\#\#\# **Example 1**

**Input Question:**

A train travels 60 kilometers in 2 hours. What is its average speed?

**Input Answer:**

The train travels a distance of 60 kilometers in 2 hours. Average speed is calculated as distance divided by time. Average speed = 60 / 2 = 30 kilometers per hour.  

**Output JSON:**
\begin{lstlisting}
{
  "templatized_question": "A train travels {distance} kilometers in {time} hours. What is its average speed?",
  "templatized_answer": [
    "The train travels a distance of {distance} kilometers in {time} hours.",
    "Average speed is calculated as distance divided by time.",
    "Average speed = {distance} / {time} = {average_speed} kilometers per hour."
  ],
  "factual_assignment": {
    "distance": 60,
    "time": 2,
    "average_speed": 30
  },

  "node_explanation":{
    "distance": "The distance traveled by the train",
    "time": "time taken by the train to travel the distance",
    "average_speed": "The average speed of the train"
  }
}
\end{lstlisting}

---

\#\#\#\# **Example 2**
**Input Question:**

Mary buys 3 books for \$15 each. How much does Mary spend in total?

**Input Answer:**

Mary buys 3 books, each costing \$15. Total cost is calculated as number of books multiplied by the cost per book. Total cost = 3 * 15 = \$45.  

**Output JSON:**

\begin{lstlisting}
{
  "templatized_question": "Mary buys {quantity} books for ${cost_per_book} each. How much does she spend in total?",
  "templatized_answer": [
    "Mary buys {quantity} books, each costing {cost_per_book}.",
    "Total cost is calculated as number of books multiplied by the cost per book.",
    "Total cost = {quantity} * {cost_per_book} = ${total_cost}."
  ],
  "factual_assignment": {
    "quantity": 3,
    "cost_per_book": 15,
    "total_cost": 45
  },
  "node_explanation":{
    "quantity": "The number of books bought by Mary",
    "cost_per_book": "The cost of each book",
    "total_cost": "The total amount spent by Mary"
  }
}
\end{lstlisting}

---

\textbf{User Prompt:} \\
f"Question: \textless question\textgreater \textbackslash nAnswer: \textless answer\textgreater \textbackslash n\textbackslash nProvide the templatized version as per the example above."

\end{tcolorbox}

\begin{tcolorbox}[colback=blue!5!white, colframe=blue!75!black, title=Code Generation LLM Prompt, breakable]

\textbf{System Prompt:} \\
Generate Python code that solves the following problem step by step:

\textbf{User Prompt:} \\

\begin{lstlisting}
Question:
<question>

Answer:
<CoT Answer>. The code must follow the variable names similar to ones in <templatized_answer>

Python Code:
\end{lstlisting}
\end{tcolorbox}

\begin{tcolorbox}[colback=blue!5!white, colframe=blue!75!black, title=Variable Extractor LLM Prompt, breakable]

\textbf{Instructions:} \\
You are an \textbf{expert in comprehension and variable extraction}. Your task is to analyze a \textbf{question}, a \textbf{step-by-step solution}, and a \textbf{dictionary of variables} and return a JSON object that adheres to the following rules:

---

\textbf{Guidelines:}
\begin{enumerate}
    \item \textbf{Inputs:}
    \begin{itemize}
        \item \textbf{Question}: The problem description.
        \item \textbf{Step-by-step solution}: The solution text, where variables may be explicitly stated or calculated.
        \item \textbf{Dictionary of variables}: Contains variable names and their descriptions. Not all variables may appear in the question or solution.
    \end{itemize}
    
    \item \textbf{Output Format:}
    \begin{itemize}
        \item Return a \textbf{JSON object} with:
        \begin{itemize}
            \item \textbf{Keys}: Variable names from the dictionary.
            \item \textbf{Values}: Numeric values extracted from the solution or question.
            \item If a value is explicitly mentioned in the \textbf{step-by-step solution}, extract it without recalculating.
            \item If the variable is not present in the solution or question, return \texttt{"None"}.
            \item Values must \textbf{preserve their original format} (e.g., fractions, decimals, or expressions).
        \end{itemize}
    \end{itemize}
    
    \item \textbf{Output Structure:}
    \begin{itemize}
        \item Enclose the output JSON object within \texttt{<JSON>} and \texttt{</JSON>} tags.
        \item All numeric values must be \textbf{string representations} (e.g., \texttt{"3/2"}, \texttt{"25.5"}, or \texttt{"12+8"}).
    \end{itemize}

    \item \textbf{Restrictions:}
    \begin{itemize}
        \item \textbf{Do not solve} the problem yourself or calculate missing values.
        \item Extract only the values as they appear in the solution.
    \end{itemize}
\end{enumerate}

\textbf{Example 1:}
\begin{lstlisting}
Question:
At a flea market, Hillary sells handmade crafts for 15 dollars per craft. 
Today, Hillary sells 6 crafts and is given an extra 5 dollars from an appreciative customer. 
Later on, Hillary deposits 12 dollars from today's profits into her bank account. 
How many dollars is Hillary left with after making the deposit?

Step-by-step solution:
Hillary earns \(15 \times 6 = 90\) dollars from selling crafts. 
Adding the extra 5 dollars, she has 90 + 5 = 95 dollars. 
After depositing 12 dollars, she has 95 - 12 = 83 dollars left.

Dictionary of variables:
{
    "price_per_craft": "The price of each craft",
    "number_of_crafts": "The number of crafts sold",
    "extra_dollars": "The extra amount given by the customer",
    "deposit_amount": "The amount deposited into the bank account",
    "total_earnings": "The total amount earned from selling crafts",
    "total_amount": "The total amount after receiving the extra dollars",
    "amount_left": "The amount left after depositing"
}

Output:
<JSON> {
    "price_per_craft": "15",
    "number_of_crafts": "6",
    "extra_dollars": "5",
    "deposit_amount": "12",
    "total_earnings": "90",
    "total_amount": "95",
    "amount_left": "83"
} </JSON>
\end{lstlisting}

\textbf{Example 2:}
\begin{lstlisting}
Question:
In a truck, there are 5 pink hard hats, 16 green hard hats, and 15 yellow hard hats. 
Carl takes away 10 pink hard hats. 
John takes away 7 pink hard hats and twice as many green hard hats as the number of pink hard hats he removed. 
Calculate the total number of hard hats that remained in the truck.

Step-by-step solution:
The total number of hats is 5 + 16 + 15 = 36. 
Carl removes 10 pink hats, leaving 36 - 10 = 26. 
John removes 7 pink hats, leaving 26 - 7 = 19. 
John also removes \(7 \times 2 = 14\) green hats, leaving \(19 - 14 = 5\) hats in total.

Dictionary of variables:
{
    "pink": "The number of pink hard hats",
    "green": "The number of green hard hats",
    "yellow": "The number of yellow hard hats",
    "carl_pink": "The number of pink hard hats taken by Carl",
    "john_pink": "The number of pink hard hats taken by John",
    "total_initial": "The total number of hats initially",
    "total_after_carl": "The total number of hats after Carl's removal",
    "total_after_john_pink": "The total number of hats after John's pink hat removal",
    "john_green": "The number of green hats taken by John",
    "total_final": "The total number of hats remaining"
}

Output:
<JSON> {
    "pink": "5",
    "green": "16",
    "yellow": "15",
    "carl_pink": "10",
    "john_pink": "7",
    "total_initial": "36",
    "total_after_carl": "26",
    "total_after_john_pink": "19",
    "john_green": "14",
    "total_final": "5"
} </JSON>
\end{lstlisting}

\textbf{User Prompt:} \\
\begin{lstlisting}
Here is the question and the step-by-step solution to the problem:

Question: {generation[6]}

Step-by-step solution: {generation[1]}

Dictionary of variables: {generation[5]}

For more detailed explanation of variables you can see how they were used in this template: {generation[2]}
\end{lstlisting}

\end{tcolorbox}
\subsection{Artifacts SynDeduct}
\label{subsec:synArt}

\begin{tcolorbox}[colback=blue!5!white, colframe=blue!75!black, title=Rule Set for SynDeduct, breakable]
\begin{lstlisting}
{
  "add": {
    "function": "lambda x, y: x + y",
    "verbalization": "{child} is the sum of {parent1} and {parent2}."
  },
  "subtract": {
    "function": "lambda x, y: x - y",
    "verbalization": "{child} is the difference between {parent1} and {parent2}."
  }
}
\end{lstlisting}
\end{tcolorbox}

\begin{tcolorbox}[colback=blue!5!white, colframe=blue!75!black, title=Data Generation Steps, breakable]
\textit{Initially, Directed Acyclic Graphs are generated. A computation graph is then constructed by selecting a fixed-length path within each DAG and randomly assigning values and operators to its nodes. A rule set, in conjunction with a predetermined collection of nouns, is employed to generate verbalization.}

\textit{Additionally, Chain-of-Thought solutions along with final answer are produced and later used to create prefixes. It is important to note that the questions generated in this process consist exclusively of "n" hops and do not include any prefix.}
\begin{lstlisting}
python3 init.py --num_graphs 99000 --m 60 --unary_ratio 0.0 --logic_mode bodmas --naming_mode noun --nouns.json --operators_file ruleset.json --output_file output.json --max_hops 24 --max_graphs 4000
 
Steps kept and undersampled to 4000:
Hop 1: 4000
Hop 2: 4000
Hop 3: 4000
Hop 4: 4000
Hop 5: 4000
Hop 6: 4000
Hop 7: 4000
Hop 8: 4000
Hop 9: 4000
Hop 10: 4000
Hop 11: 4000
Hop 12: 4000
Hop 13: 4000
Hop 14: 4000
Hop 15: 4000
Hop 16: 4000
Hop 17: 4000
Hop 18: 4000
Hop 19: 4000
Hop 20: 4000
Hop 21: 4000
Hop 22: 4000
Hop 23: 4000
Hop 24: 4000
\end{lstlisting}
\textit{The chain-of-solution is now appended to the question, resulting in questions that incorporate a specified number of prefixes alongside n hops. To create a balanced dataset, the maximum number of hops is limited to 12, half the total hops, and the total number of items is capped at 120.}

\textit{For instance, in the case of Hop12, there are 10 questions featuring a 12-hop prefix. This configuration implies that each such question originated from a 24-hop question, wherein the first 12 hops, serving as the prefix of the chain-of-thought, are provided, and the language model is required to resolve the remaining 12 hops.}
\begin{lstlisting}
python transformer.py --max_hops 12 --max_items 120 --max_prefixes 12 --max_prefix_length 10 output.json

Prefix Length Distribution Per Hop Category (After Undersampling):

Hop 1: Prefix1: 10, Prefix2: 10, Prefix3: 10, Prefix4: 10, Prefix5: 10, Prefix6: 10, Prefix7: 10, Prefix8: 10, Prefix9: 10, Prefix10: 10, Prefix11: 10, Prefix12: 10
Hop 2: Prefix1: 10, Prefix2: 10, Prefix3: 10, Prefix4: 10, Prefix5: 10, Prefix6: 10, Prefix7: 10, Prefix8: 10, Prefix9: 10, Prefix10: 10, Prefix11: 10, Prefix12: 10
Hop 3: Prefix1: 10, Prefix2: 10, Prefix3: 10, Prefix4: 10, Prefix5: 10, Prefix6: 10, Prefix7: 10, Prefix8: 10, Prefix9: 10, Prefix10: 10, Prefix11: 10, Prefix12: 10
Hop 4: Prefix1: 10, Prefix2: 10, Prefix3: 10, Prefix4: 10, Prefix5: 10, Prefix6: 10, Prefix7: 10, Prefix8: 10, Prefix9: 10, Prefix10: 10, Prefix11: 10, Prefix12: 10
Hop 5: Prefix1: 10, Prefix2: 10, Prefix3: 10, Prefix4: 10, Prefix5: 10, Prefix6: 10, Prefix7: 10, Prefix8: 10, Prefix9: 10, Prefix10: 10, Prefix11: 10, Prefix12: 10
Hop 6: Prefix1: 10, Prefix2: 10, Prefix3: 10, Prefix4: 10, Prefix5: 10, Prefix6: 10, Prefix7: 10, Prefix8: 10, Prefix9: 10, Prefix10: 10, Prefix11: 10, Prefix12: 10
Hop 7: Prefix1: 10, Prefix2: 10, Prefix3: 10, Prefix4: 10, Prefix5: 10, Prefix6: 10, Prefix7: 10, Prefix8: 10, Prefix9: 10, Prefix10: 10, Prefix11: 10, Prefix12: 10
Hop 8: Prefix1: 10, Prefix2: 10, Prefix3: 10, Prefix4: 10, Prefix5: 10, Prefix6: 10, Prefix7: 10, Prefix8: 10, Prefix9: 10, Prefix10: 10, Prefix11: 10, Prefix12: 10
Hop 9: Prefix1: 10, Prefix2: 10, Prefix3: 10, Prefix4: 10, Prefix5: 10, Prefix6: 10, Prefix7: 10, Prefix8: 10, Prefix9: 10, Prefix10: 10, Prefix11: 10, Prefix12: 10
Hop 10: Prefix1: 10, Prefix2: 10, Prefix3: 10, Prefix4: 10, Prefix5: 10, Prefix6: 10, Prefix7: 10, Prefix8: 10, Prefix9: 10, Prefix10: 10, Prefix11: 10, Prefix12: 10
Hop 11: Prefix1: 10, Prefix2: 10, Prefix3: 10, Prefix4: 10, Prefix5: 10, Prefix6: 10, Prefix7: 10, Prefix8: 10, Prefix9: 10, Prefix10: 10, Prefix11: 10, Prefix12: 10
Hop 12: Prefix1: 10, Prefix2: 10, Prefix3: 10, Prefix4: 10, Prefix5: 10, Prefix6: 10, Prefix7: 10, Prefix8: 10, Prefix9: 10, Prefix10: 10, Prefix11: 10, Prefix12: 10

Total Prefix Length Distribution Across Hops:

Prefix1: 120
Prefix2: 120
Prefix3: 120
Prefix4: 120
Prefix5: 120
Prefix6: 120
Prefix7: 120
Prefix8: 120
Prefix9: 120
Prefix10: 120
Prefix11: 120
Prefix12: 120

Number Of Items per Hop

Hop: 1 - 120
Hop: 2 - 120
Hop: 3 - 120
Hop: 4 - 120
Hop: 5 - 120
Hop: 6 - 120
Hop: 7 - 120
Hop: 8 - 120
Hop: 9 - 120
Hop: 10 - 120
Hop: 11 - 120
Hop: 12 - 120
Total entries in transformed JSON: 1440
\end{lstlisting}
\end{tcolorbox}

\begin{tcolorbox}[colback=blue!5!white, colframe=blue!75!black, title=A single Data-point of SynDeduct, breakable]
\textbf{Prompt Part A: Graph Structure and Question (will be given as user)}
\textit{The graph structure encompasses the complete verbalization of the entire graph, whereas the question is derived solely from a specific path within that graph. Consequently, a considerable amount of the information contained in the graph structure is not necessary for generating a solution. This design serves to assess the model's capability to extract and utilize only the relevant information from a broader context.}
\begin{lstlisting}
=== Graph Structure ===
Inputs:
  - Masako (value = 8)
  - Nalca (value = 2)
  - Gassman (value = 5)
  Derived Nodes:
  - Certain is the sum of Nalca and Masako.
  - Irtysh is the sum of Certain and Gassman.
  - Horstman is the difference between Masako and Certain.
  - Pellicano is the difference between Horstman and Gassman.
  - Taoiseach is the difference between Masako and Gassman.
  - Vanvalkenburg is the difference between Gassman and Certain.
  - Nourse is the sum of Irtysh and Nalca.
  - Clapham is the sum of Pellicano and Taoiseach.
  - Nuncio is the difference between Nalca and Horstman.
  - Foxbat is the difference between Nalca and Gassman.
  - Kenyon is the sum of Nuncio and Masako.
  - Riva is the sum of Kenyon and Nourse.
  - Claymore is the difference between Irtysh and Riva.
  - Ballville is the sum of Masako and Riva.
  - Lai is the difference between Kenyon and Clapham.
  - Smolik is the sum of Vanvalkenburg and Riva.
  - Bushi is the sum of Horstman and Claymore.
  - Batiste is the sum of Riva and Kenyon.
  - Criner is the sum of Riva and Certain.
  - Begnaud is the difference between Nourse and Foxbat.
  - SEPA is the sum of Certain and Irtysh.
  - Wentling is the sum of Nalca and Smolik.
  - Troon is the sum of Lai and Begnaud.
  - Sanderson is the sum of Wentling and Begnaud.
  - Ferozepore is the difference between Horstman and Sanderson.
  - Sibiu is the sum of Ballville and Riva.
  - Bootle is the sum of Irtysh and Nalca.
  - Climategate is the sum of Vanvalkenburg and Taoiseach.
  - Maland is the difference between Certain and Vanvalkenburg.
  - Hobby is the difference between Sanderson and Kenyon.
  - Tikrit is the difference between Nourse and Bootle.
  - Lamarca is the sum of Maland and Criner.
  - Dnipr is the sum of Irtysh and Nourse.
  - Arvid is the difference between SEPA and Horstman.
  - Plath is the sum of SEPA and Criner.
  - Gulliver is the difference between Kenyon and Sibiu.
  - Helatrobus is the difference between Plath and Sanderson.
  - Tulu is the sum of Nalca and Kenyon.
  - Shuka is the sum of Nourse and Vanvalkenburg.
  - Hemsley is the difference between Bootle and Pellicano.
  - Creasman is the sum of Nourse and Troon.
  - Falcon is the sum of Clapham and Irtysh.
  - Border is the difference between Gassman and Tikrit.
  - Noyola is the difference between Lamarca and Hobby.
  - Tommie is the sum of Taoiseach and Helatrobus.
  - Hines is the sum of Masako and Batiste.
  - Adney is the difference between Wentling and Bushi.
  - Winsford is the sum of Ballville and Shuka.
  - Iga is the sum of Plath and Riva.
  - Jacqueline is the sum of Winsford and Vanvalkenburg.
  - Wheatley is the sum of Border and Irtysh.
  - Lyndon is the sum of Noyola and Shuka.
  - Arvelo is the difference between Pellicano and SEPA.
  - Belphegor is the difference between Helatrobus and Wheatley.
  - Kassandra is the difference between Bootle and Tulu.
  - Garth is the difference between Wentling and Plath.
  - Yucatec is the sum of Pellicano and Hines.

  What is the value of Arvelo?
\end{lstlisting}
\textbf{Prompt PartB: Partial COT (prefix)}

\textit{This assistant prompt-response pair will be provided to the language model with the eos\_token removed from the end. This ensures that the model continues generating text seamlessly from where the given prefix ends, thereby guiding its output to align with the intended structure and constraints.}
\begin{lstlisting}
Answer: - Given value of Masako = 8 .
- Given value of Nalca = 2 .Certain is the sum of Nalca and Masako.
The Computed value of Certain = 10 
\end{lstlisting}
\textbf{Response From LLM}
\begin{lstlisting}
    - Given value of Gassman = 5 .  Irtysh is the sum of Certain and Gassman. The Computed value of Irtysh = 15  Horstman is the difference between Masako and Certain. The Computed value of Horstman = -2  Pellicano is the difference between Horstman and Gassman. The Computed value of Pellicano = -7  Taoiseach is the difference between Masako and Gassman. The Computed value of Taoiseach = 3  Vanvalkenburg is the difference between Gassman and Certain. The Computed value of Vanvalkenburg = -5  Nourse is the sum of Irtysh and Nalca. The Computed value of Nourse = 15 + 10 = 15  The Computed value of Nourse = 100 + 10 = 100  The Computed value of Nourse = 100 + 10 = 110 + 15 = 115  The Computed value of Nourse = 15 + 10 = 110 + 15 = 115  The Computed value of Nourse = 100 + 15 = 115  The Computed value of Nourse = 15  The Computed value of Nourse = 15  The Computed value of Nourse = 15 + 10 = 25  The Compu
ted value of Nourse = 100  The Computed value of Nourse = 100 + 15 = 115  The Computed value of Nourse = 15 + 15 = 30  The Computed value of Nourse = 100  T
he Computed value of Nourse = 100  The value of Septembr = 100  The value of Rondeau = 10  The value of Rondeau = 10  The value of Rondeau = 10  The value o
f Nalca = 10  The value of Nalca = 10  The value of Nalca = 10  The value of Nalca = 10  The value of Nalca = 10  The value of Nalca = 10  The value of Nalc
a = 10  The value of Nalca = 10  The value of Nalca = 10  The value of Nalca = 10  The value of Nalca = 10  The value of Nalca = 10  The value of Nalca = 10
  The value of Nalca = 10  The value of Nalca = 10  The value of Malca = 10  The value of Malca = 10  The value of Malca = 10  The value of Maland = 10  The
 value of Maland = 10  The value of Mand = 100  The value of Mand = 100  The value of Nalca = 100  The value of Nalca = 100  The value of Nalca = 10  The va
lue of Salca = 10  The value of Valca = 10  The value of Calca = 10  The value of Galca = 10  The value of Palca = 10  The value of Balca = 10  The value of
 Talca = 10  The value of Falca = 10  SEPA is the sum of Certain and Irtysh. The Computed value of SEPA = 10 + 15 = 25  Pellicano is the difference between
Horstman and Gassman. The Computed value of Pellicano = -2 - 5 = -7  Arvelo is the difference between Pellicano and SEPA. The Computed value of Arvelo = -7
- 25 = -32 \n#### -3
\end{lstlisting}
\textbf{Template System Prompt}
\begin{lstlisting}
You are a computation graph reasoning assistant designed to evaluate mathematical expressions described in any style of verbalizations. Your task is to process a graph structure, interpret the relationships between nodes based on the provided verbalizations, and answer questions about specific nodes. 

Here are the rules and expectations for your behavior: 
--- 
### Rules: 
{RuleSet.json is filled here}

**Graph Structure Processing**: 
   - Nodes are defined as inputs or derived nodes. 
   - Inputs have predefined values. 
   - Derived nodes depend on other nodes and their relationships as defined by verbalizations. 

**Step-by-Step Reasoning**: 
   - Interpret the graph structure line-by-line. 
   - Calculate the value of each derived node based on its dependencies, ensuring that the verbalization is correctly mapped to its mathematical function. 
   - Use previously calculated or input values as required. 

**Answer Presentation**: 
   - Provide the value of the requested node only after completing all necessary computations and make sure the value is a integer or a float. 
   - SHOW THE REASONING STEP-BY-STEP AND PROVIDE THE FINAL ANSWER CLEARLY, PREFIXED BY `####` and NOTHING AFTER IT. 
   - Suppose answer is 56. You must output `#### 56` at the end of each step-by-step solution. 

   Example 1: 

	{Graph Structure is filled here}  
	{Question is filled here}

  	Answer:  Rondeau is an input with value 10.
			 - Septembr is the square of Rondeau.. The value of Septembr = 100 
			 ####100
\end{lstlisting}
\end{tcolorbox}

\textit{Subsequently, the output generated by the language model is processed using a Variable Extractor analogous to that employed in the GSM8K dataset. The parsed response is then normalized—massaged into the correct format (for instance, converting fractional representations to floating-point numbers)—and subsequently compared to the final expected answer, allowing for a tolerance of up to 5 per-cent deviation from the original value.}